\PassOptionsToPackage{colorlinks=true,linkcolor=blue,citecolor=blue,urlcolor=blue}{hyperref}
    \documentclass[lettersize,journal]{IEEEtran}

\usepackage{amsmath,amsfonts,amssymb,bm}
\usepackage{array}
\usepackage{booktabs}
\usepackage{cite}           
\usepackage{graphicx}
\usepackage{multirow}
\usepackage{pifont}
\usepackage[caption=false,font=normalsize,labelfont=sf,textfont=sf]{subfig}
\usepackage{stfloats}
\usepackage{textcomp}
\usepackage{orcidlink} % 加载orcid图标宏包
\usepackage{url}
\usepackage[table,dvipsnames]{xcolor}
\usepackage[colorlinks=true,linkcolor=blue,citecolor=blue,urlcolor=blue]{hyperref}
\usepackage{booktabs}
\usepackage{multirow}
\usepackage{graphicx}

\graphicspath{{./}}
\begin{document}
\bstctlcite{IEEEexample:BSTcontrol}
\title{Training-Free Interaction-Aligned Visual Token Pruning for Efficient Embodied Manipulation}

\author{Jintao Cheng\orcidlink{0009-0008-0121-4162},
        Weibin Li,
        Haozhe Wang\orcidlink{0009-0006-2854-7464},
        Gang Wang,
        Yipu Zhang\orcidlink{0009-0007-9465-5807},
        Xiaoyu Tang\orcidlink{0000-0002-6038-9623},
        Jin Wu\orcidlink{0000-0001-5930-4170},\\
        Xieyuanli Chen\orcidlink{0000-0003-0955-6681},
        Yunhui Liu,~\IEEEmembership{Fellow,~IEEE}\orcidlink{0000-0002-7297-1493},
        and Wei Zhang,~\IEEEmembership{Fellow,~IEEE}\orcidlink{0000-0002-7622-6714}%

\thanks{Manuscript received X X, 20XX; revised X X, 20XX; accepted X X, 20XX. (Corresponding author: Wei Zhang.)}%
\thanks{J. Cheng, Y. Zhang, and W. Zhang are with the Department of Electronic and Computer Engineering, The Hong Kong University of Science and Technology, Hong Kong, China (e-mail: \texttt{jchengau@connect.ust.hk, yzhangqg@connect.ust.hk, eeweiz@ust.hk}).}%
\thanks{W. Li is with the University of Macau, Macau, China (e-mail: \texttt{liweibin0830@gmail.com}).}%
\thanks{H. Wang and X. Tang are with the School of Data Science and Engineering, Xingzhi College, South China Normal University, Guangzhou, China (e-mail:\texttt{202481324245@m.scnu.edu.cn, tangxy@scnu.edu.cn}).}%
% \thanks{W. Li is with the University of Macau, Macau, China (e-mail: \texttt{liweibin0830@gmail.com}).}%
\thanks{G. Wang and Y. Liu are with the Department of Mechanical and Automation Engineering, The Chinese University of Hong Kong, Hong Kong, China (e-mail: \texttt{gwang2@mae.cuhk.edu.hk, yhliu@cuhk.edu.hk}).}%
% \thanks{X. Tang is with the School of Electronics Science and Engineering and Xingzhi College, South China Normal University, Guangzhou, China (e-mail: \texttt{tangxy@scnu.edu.cn}).}%
\thanks{J. Wu is with the School of Intelligence Science and Technology, University of Science and Technology Beijing, Beijing, China (e-mail: \texttt{wujin@ustb.edu.cn}).}%
\thanks{Xieyuanli Chen is with the College of Intelligence Science and Technology, National University of Defense Technology, Changsha 410073, China (e-mail: \texttt{xieyuanli.chen@nudt.edu.cn}).}%
% \thanks{Y. Liu is with The Chinese University of Hong Kong, Hong Kong, China, and also with the Hong Kong Centre for Logistics Robotics, Hong Kong, China (e-mail: \texttt{yhliu@cuhk.edu.hk}).}%

}

\markboth{IEEE Transactions on Image Processing}%
{Cheng \MakeLowercase{\textit{et al.}}: IAprune: Training-Free Visual Token Pruning via Interaction Alignment}

\maketitle

\begin{abstract}
Efficient visual representation is a central image-processing challenge in embodied manipulation, where policies repeatedly process dense visual-token sequences during closed-loop control. Existing methods rank or prune tokens using semantic relevance, VLM attention, cross-frame redundancy, or motion in the action space. These signals may discard task-relevant regions when instruction-related appearance and observed image motion are not yet spatially aligned. We introduce Interaction-Aligned Pruning (IAprune), a training-free method that treats per-frame budget setting and within-budget token selection as two linked decisions. Semantic--motion spatial agreement guides the choice between Conservative and Aggressive coverage, and the resulting region size is mapped to a calibrated dynamic budget. Continuous semantic and motion responses rank the existing tokens, while geometric residual correction redirects fixed selection slots toward under-represented boundaries without increasing the sequence length. Across four embodied manipulation policies, three simulation benchmarks, and a real-robot platform, IAprune provides a favorable accuracy--efficiency trade-off, matching the unpruned policy on LIBERO with a \(1.54\times\) speedup and reaching \(1.48\times\) acceleration on a real robot. Phase-wise analysis shows that retention gains are largest under tight budgets early in an episode, while fixed-budget analysis confirms that geometry replaces low-priority tokens with boundary and contact-region evidence rather than retaining more tokens. Our project website is: \href{https://chengjt1999.github.io/VLA-IAP.github.io/}{IAprune.com}.

\end{abstract}

\begin{IEEEkeywords}
Visual token pruning, embodied manipulation, efficient inference.
\end{IEEEkeywords}

\section{Introduction}
\label{sec:intro}
\begin{figure*}[!t]
    \centering
    \includegraphics[width=\textwidth]{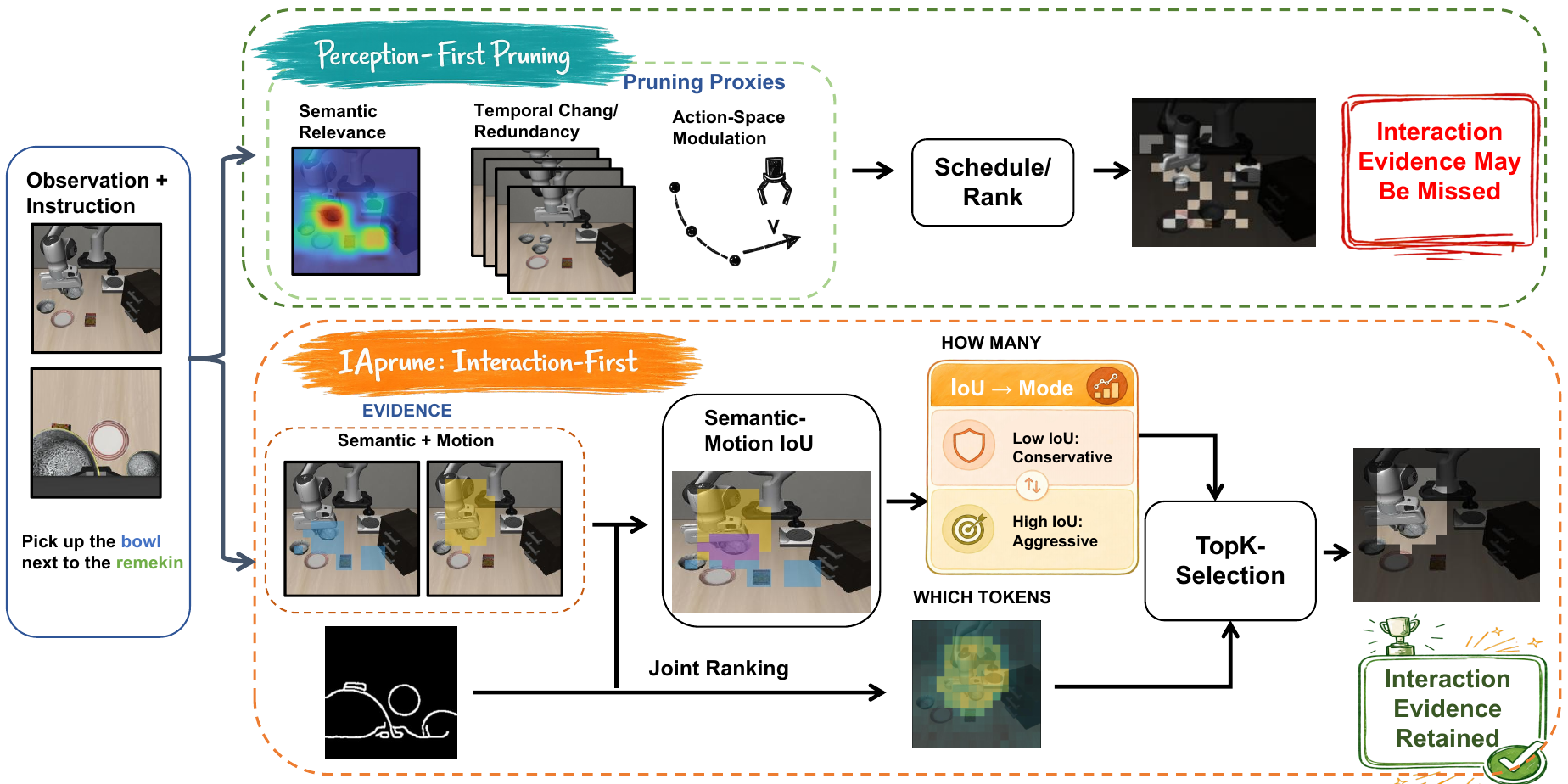}
    \caption{\textbf{Comparison of Perception-First vs. Interaction-First token pruning paradigms.} Perception-First baselines (Top) prematurely lose the manipulation target due to early semantic misalignment---a vulnerability that simple temporal stacking fails to resolve without explicit interaction modeling. In contrast, IAprune (Bottom) combines IoU-aware conservative-to-aggressive scheduling with budget-preserving structural reallocation, preserving the physical target while reducing redundant visual computation.}
    \label{fig:motivation_comparison}
\end{figure*}

\IEEEPARstart{E}{fficient} visual representation is becoming a central image-processing problem in embodied manipulation, where policies repeatedly convert high-resolution observations into dense token sequences for closed-loop control. This processing cost is shared by transformer-based controllers and large pretrained policies, including RT-1~\cite{brohan2022rt}, OpenVLA~\cite{kim2024openvla}, and $\pi_0$~\cite{black2024pi_0}. It also arises in recent models that augment action prediction with future-image or video-state prediction~\cite{cen2025worldvla,ye2026dreamzero,liao2026genie}. Despite their different designs, these models must encode and process visual tokens at every control step. The cost grows with image resolution, camera count, and temporal context. Efficient visual-token processing is therefore a general image-processing challenge for embodied manipulation, not an optimization tied to a specific policy family.

Visual-token pruning can reduce this cost without changing the underlying policy. For vision-language model (VLM) inference, existing methods rank tokens using early-layer attention, as in FastV~\cite{chen2024image}; prompt-conditioned text--vision relevance, as in SparseVLM~\cite{zhang2024sparsevlm}; or feature coverage and diversity, as in DivPrune~\cite{alvar2025divprune}. These signals are designed to remove redundant visual content in image and video understanding. Embodied manipulation, however, requires the retained tokens to support control over time. The relevant visual evidence changes as an episode unfolds. A patch with a weak semantic response may still capture a thin handle, an object boundary, or a region near contact that is needed for precise motion.

Recent methods for accelerating embodied policies have started to account for this temporal variation. VLA-Cache~\cite{xu2025vla} reuses stable visual-token states and recomputes regions selected by decoder attention. EfficientVLA~\cite{yang2025efficientvla} combines attention-based task relevance with feature diversity in a structured acceleration framework. SpecPrune-VLA~\cite{wang2025specprune} adds changing image patches to historical backbone attention and adapts pruning based on end-effector velocity. VLA-ADP~\cite{pei2025action} instead gates text-guided pruning using recent action trajectories. Together, these methods treat token compression as a dynamic process rather than a fixed decision. Their control signals, however, mainly describe visual redundancy, model responses, or motion granularity in the action space. They do not directly measure whether task-relevant appearance and observed image motion agree on the same spatial region. They also do not use independent structural cues to recover low-saliency boundaries under a fixed token budget.

This gap is most important early in an interaction, before the visual evidence becomes stable. At the start of a manipulation episode, the target may be off-center, partly occluded, or surrounded by clutter. Motion cues may also be weak or distorted by camera and background changes. A saliency-based selector may then focus on the wrong region. A controller driven by action magnitude can adapt compression to the motion phase, but it cannot confirm that task-relevant appearance and observed image motion are spatially aligned. Even after the target is located, low-texture edges and narrow contact regions may receive low scores. As illustrated in Fig.~\ref{fig:motivation_comparison}, token pruning for manipulation must therefore address two coupled questions. \emph{How many tokens should be retained?} The budget should adapt to the current interaction certainty. \emph{Which tokens should be retained within that budget?} Their selection should combine semantic, motion, and geometric evidence.

We refer to this design as \textbf{Interaction-First visual-token pruning}. It uses the spatial agreement between task-conditioned semantic responses and observed image motion as a practical cue for adjusting compression. Following this idea, we develop \textbf{Interaction-Aligned Pruning} (IAprune), a training-free method with three steps: measuring semantic--motion agreement, setting a per-frame token budget, and selecting tokens within that budget. Semantic--motion IoU selects between Conservative and Aggressive coverage, and target-average calibration maps the resulting region size to the budget \(K_t\). The continuous semantic and motion responses provide the initial token ranking. Geometry then acts as a residual cue and reallocates a fixed number of selection slots toward informative boundaries that the initial ranking may miss. In this way, semantic--motion agreement guides the budget, while semantic, motion, and geometry jointly determine which existing tokens are retained.

We evaluate IAprune across multiple embodied manipulation policies, three simulation benchmarks, and a real-robot platform. On LIBERO~\cite{liu2023libero}, IAprune matches the unpruned policy's 97.1\% average success rate at the target 30\% setting while providing a $1.54\times$ speedup; real-robot tests report up to $1.48\times$ acceleration. Beyond these headline results, controlled analyses examine when interaction alignment is most useful. Under the 30\% budget, early-stage retention of target, boundary, and contact-region evidence reaches 0.733/0.672/0.710, compared with 0.374/0.311/0.313 for the strongest baseline. This gap narrows at 70\% retention, suggesting that alignment-guided scheduling matters most when visual evidence is uncertain and the token budget is tight. Fixed-budget analysis further shows that geometric residual correction replaces low-priority tokens with boundary and contact-region evidence rather than retaining more tokens. Together, these results support a scoped interpretation of the method: alignment-guided scheduling protects uncertain phases, while structural reallocation helps when sparse geometric cues would otherwise receive low scores.

The main contributions are summarized as follows:

1. We present an \textbf{Interaction-First} formulation for visual-token pruning in embodied manipulation. It treats per-frame budget setting and within-budget token selection as two linked image-processing decisions. Semantic--motion agreement guides the budget, while continuous semantic and motion responses rank the tokens to retain.

\par
2. We develop \textbf{Budget-Preserving Structural Reallocation}, which uses lightweight geometry as a residual cue to redirect fixed selection slots toward under-represented boundaries. The module changes which existing tokens are retained without modifying token features or the scheduler-assigned budget.

\par
3. We evaluate IAprune across multiple embodied manipulation policies, three simulation benchmarks, and real-robot tasks. Main comparisons measure the accuracy--efficiency trade-off, while component, phase-wise retention, fixed-budget allocation, robustness, latency, and memory analyses characterize when each design choice is useful. The results suggest that alignment-guided scheduling matters most in uncertain, low-budget phases, while structural reallocation recovers sparse boundary evidence without increasing the token count. Code will be publicly released.

\section{Related Work}
\subsection{Visual Modeling for Embodied Manipulation}
Embodied manipulation systems use different policy formulations, but they all depend on visual representations extracted from image or video observations. Direct action-prediction policies map image--language context to discrete actions, autoregressive action tokens, or continuous action chunks. Representative examples include RT-1~\cite{brohan2022rt}, OpenVLA~\cite{kim2024openvla}, OpenVLA-OFT~\cite{kim2025fine}, CogACT~\cite{li2024cogact}, \(\pi_0\)~\cite{black2024pi_0}, and \(\pi_{0.5}\)~\cite{intelligence2025pi05}. Other systems add visual prediction to the control process: WorldVLA~\cite{cen2025worldvla} jointly models image and action tokens, while DreamZero~\cite{ye2026dreamzero} and Genie Envisioner~\cite{liao2026genie} predict or transfer future visual representations for action generation. Complementing these policy-level designs, KIM~\cite{xuan2024knowledge} injects task-agnostic knowledge into visual features, OATNet~\cite{yang2022oatnet} jointly encodes text and pixel tokens, and HiSA~\cite{xu2022hisa} aligns language queries with temporally localized video representations. Across these policy architectures, visual representations are repeatedly processed during closed-loop execution. IAprune targets this shared image-processing cost at inference time.

Manipulation also requires visual evidence beyond generic semantic saliency. Classical visual manipulation policies encode spatial structure through feature transport or voxelized action prediction~\cite{zeng2021transporter,shridhar2023peract}. More recent approaches ground referred or actionable image regions, affordances, and relational keypoint constraints among objects, robot parts, and target locations~\cite{liu2023mmii,11551826,huang2024rekep}. These formulations highlight the importance of handles, object boundaries, contact-adjacent surfaces, and other sparse geometric evidence. Such regions may receive weak scores from language-conditioned relevance alone. The cited methods mainly learn manipulation representations, affordances, or task constraints. IAprune instead selects visual tokens at inference time. Its semantic, motion, and geometric cues are used to preserve evidence for the downstream policy.

\subsection{Visual Token Compression for Embodied Manipulation}
Efficient visual processing in transformer models has been studied through adaptive computation, token pruning and merging, structured model reduction, and cache sparsification. In image models, computation can be reduced through adaptive feature refinement~\cite{han2021safr}, dynamic token pruning~\cite{rao2021dynamicvit}, token merging~\cite{10183862}, or efficient self-attention~\cite{xu2025s2aformer}. Methods for vision-language model (VLM) inference use lightweight architectures~\cite{luo2022lightweight}, attention- or prompt-based relevance~\cite{chen2024image,zhang2024sparsevlm}, adaptive thresholds or token transitions~\cite{ye2025atpllava,li2026transprune}, and attention-head information or feature diversity~\cite{11508157,alvar2025divprune}. Cross-modal key--value states can also be compressed~\cite{pei2024cross}. These methods identify redundant visual or cross-modal content, but most are designed for image or video understanding. They do not directly track how control-relevant visual evidence changes during closed-loop manipulation.

Recent methods for accelerating embodied policies introduce temporal and action-aware signals into compression. VLA-Cache~\cite{xu2025vla} combines cross-frame visual stability with decoder attention to reuse token states. TeamVLA~\cite{ye2025token} expands and merges tokens under action-aware guidance. EfficientVLA~\cite{yang2025efficientvla} combines task relevance, feature diversity, and caching in a structured acceleration pipeline. SP-VLA~\cite{li2025sp} jointly schedules policy execution and performs spatio-semantic token pruning, while VLA-Pruner~\cite{liu2025vla} combines semantic prefill relevance with temporally smoothed action relevance. Learned approaches provide another option by optimizing token retention through differentiable pruning~\cite{jiang2025better}. Together, these methods treat visual-token compression as a temporal and multi-signal process. They differ mainly in which signals control the budget and the retained regions.

The methods closest to IAprune use robot motion to adjust pruning. SpecPrune-VLA~\cite{wang2025specprune} combines historical backbone attention and current visual changes with an end-effector-velocity controller. VLA-ADP~\cite{pei2025action} uses recent end-effector trajectories to gate text-guided token pruning. Both methods use motion in the action space to adapt compression. IAprune instead uses the spatial agreement between instruction-relevant appearance and observed image motion as a cue for setting the visual-token budget. Semantic--motion agreement guides the per-frame budget \(K_t\), continuous semantic and motion responses rank the existing tokens, and geometry redirects fixed selection slots toward under-represented boundaries. This distinction concerns how tokens are selected under a fixed budget; it does not imply predicting physical contact or modifying token features.

\section{Methodology}
\label{sec:method}

\begin{figure*}[!t]
    \centering
    \includegraphics[width=\textwidth]{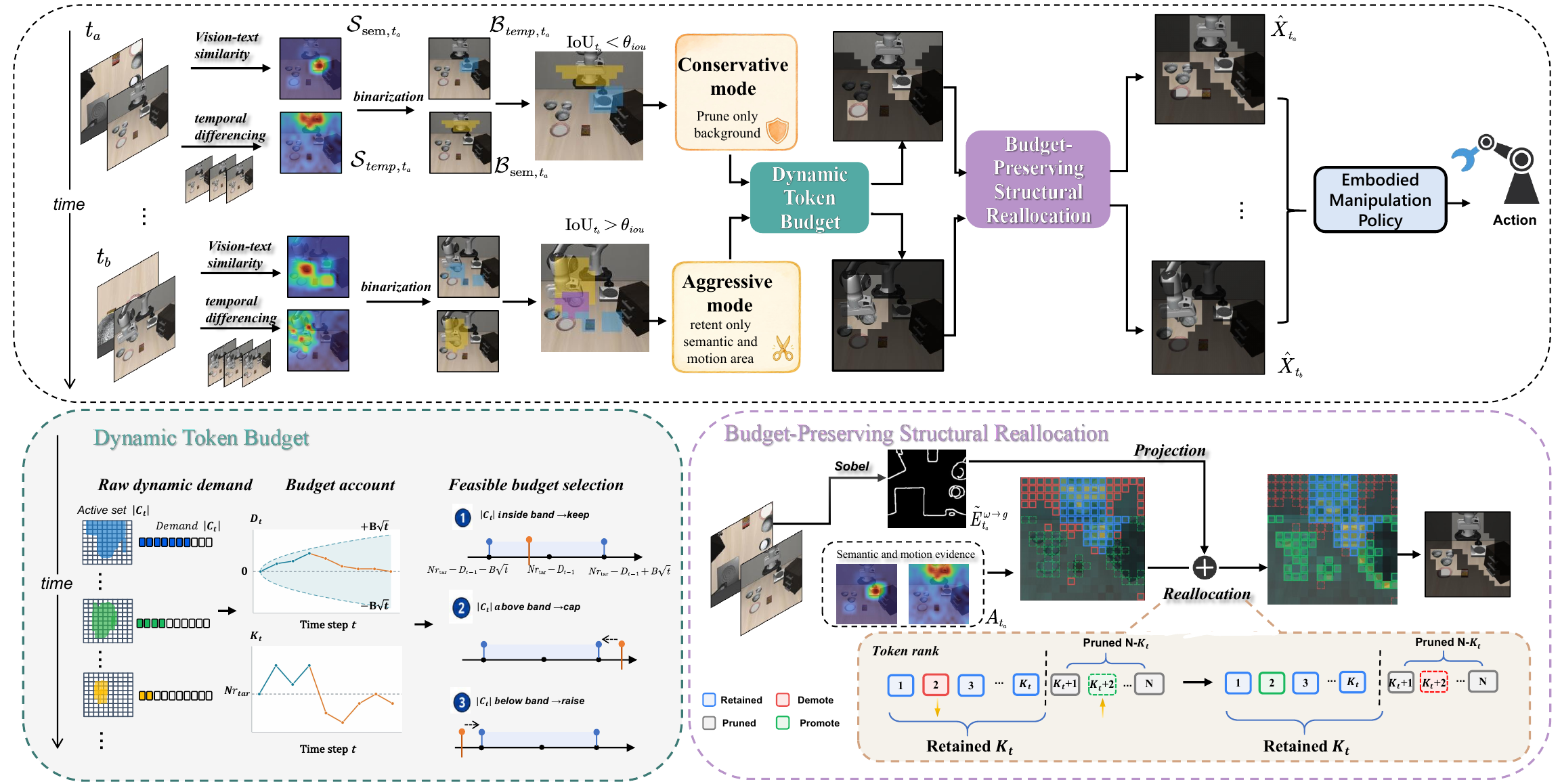}
       \caption{\textbf{Overview of IAprune.} Semantic and motion evidence serve two complementary roles: their spatial agreement selects a Conservative or Aggressive coverage mode, while their continuous responses define the base relevance of each visual token. The selected mode defines a provisional region, whose size is converted into a dynamic per-frame budget \(K_t\) under a target average retention ratio \(r_{\mathrm{tar}}\). Geometric evidence then performs budget-preserving structural reallocation through residual correction without changing \(K_t\). The resulting compact visual sequence is fed into the unchanged embodied manipulation policy for action generation.}
    \label{fig:framework}
\end{figure*}

\subsection{Overview}
IAprune treats visual-token pruning as two linked decisions: how many tokens to retain from each RGB frame and which existing tokens to keep within that budget. Semantic and motion evidence inform both decisions, while geometry is used only in the final selection stage.

As shown in Fig.~\ref{fig:framework}, semantic--motion agreement selects Conservative or Aggressive coverage, whose size is mapped to a dynamic per-frame budget \(K_t\) under the target retention ratio \(r_{\mathrm{tar}}\). Continuous semantic and motion responses rank the tokens, and geometry refines the selection within the same \(K_t\). The compact sequence is then passed to the unchanged embodied manipulation policy.

\subsection{Problem Formulation}
IAprune addresses a per-frame image-processing problem: reducing the visual-token representation of an RGB image before policy inference while retaining the information needed for control. At time \(t\), the direct visual input is an RGB image \(I_t\in\mathbb{R}^{H\times W\times3}\), paired with a language instruction \(q\). A frozen visual encoder maps the image to

\begin{equation}
X_t^{\mathrm{vis}}
=
\mathcal{E}_{\mathrm{vis}}(I_t)
=
[x_{t,1},\ldots,x_{t,N}]^\top
\in\mathbb{R}^{N\times D}.
\label{eq:input_tokens}
\end{equation}
The corresponding instruction representation is
\begin{equation}
e_{\mathrm{text}}
=
\mathcal{E}_{\mathrm{text}}(q)
\in\mathbb{R}^{D}.
\label{eq:text_feature}
\end{equation}
Here, \(N=(H/P)(W/P)\), where \(P\) is the patch size and \(D\) is the feature dimension; each token \(x_{t,i}\) is associated with spatial coordinate \(\mathbf{u}_i\). IAprune returns an ordered subsequence
\begin{equation}
\hat X_t
=
[x_{t,i_1},\ldots,x_{t,i_{K_t}}]^\top .
\label{eq:pruned_sequence}
\end{equation}
Here, \(\{i_1,\ldots,i_{K_t}\}\subseteq\{1,\ldots,N\}\), and the cardinality \(K_t\) may vary across frames. The method operates between feature extraction and policy inference and does not modify the backbone or action-generation parameters.

\subsection{Semantic--Motion Interaction Modeling}
Semantic and motion evidence describe different aspects of the current frame. Semantic evidence highlights regions relevant to the instruction, but its response may be broad or may emphasize visually recognizable context. Motion evidence highlights local scene changes, but it may also respond to the manipulator, camera variation, or task-irrelevant objects. IAprune therefore does not use either response as a standalone pruning mask. Instead, their thresholded supports measure spatial agreement and select a coverage mode, while their continuous scores rank the existing tokens. In this way, the same evidence guides both how many tokens are retained and which tokens are selected.

\subsubsection{Semantic Evidence}
We first compute the spatial mean of the visual tokens:
\begin{equation}
\bar{x}_t
=
\frac{1}{N}\sum_{j=1}^{N}x_{t,j}.
\label{eq:spatial_mean}
\end{equation}
and use it to center and normalize each visual feature:
\begin{equation}
\hat{x}_{t,i}
=
\frac{x_{t,i}-\bar{x}_t}
{\|x_{t,i}-\bar{x}_t\|_2}.
\label{eq:visual_normalization}
\end{equation}
The text feature is normalized independently:
\begin{equation}
\hat{e}_{\mathrm{text}}
=
\frac{e_{\mathrm{text}}}
{\|e_{\mathrm{text}}\|_2}.
\label{eq:text_normalization}
\end{equation}
Spatial centering reduces the component shared across all patches, while \(L_2\) normalization limits the effect of feature magnitude on cross-modal comparison. We then compute the instruction-conditioned distribution

{\small
\begin{equation}
\displaystyle
p_{t,i} =
\frac{\exp(\hat{x}_{t,i}\cdot\hat{e}_{\mathrm{text}})}
{\sum_{j=1}^{N}\exp(\hat{x}_{t,j}\cdot\hat{e}_{\mathrm{text}})}.
\end{equation}}
Because the normalized dot product measures directional agreement, \(p_{t,i}\) represents the relative compatibility between token \(i\) and the instruction within the current frame. We then apply spatial average pooling to reduce isolated peaks and spread evidence across neighboring patches. Min--max normalization produces \(\mathcal{S}_{\mathrm{sem},t}\in[0,1]^N\). This map does not directly decide which tokens to keep. Its thresholded support is used to measure semantic--motion agreement, while its continuous values remain available for token ranking.

\subsubsection{Motion Evidence}
Action-space velocity may vary sharply during early manipulation and does not directly describe where changes occur in the image. We therefore obtain motion evidence from consecutive visual features. For token \(i\), the second-order temporal difference is

\begin{equation}
d_{t,i}
=
\left\|
\hat{x}_{t,i}-2\hat{x}_{t-1,i}+\hat{x}_{t-2,i}
\right\|_2 .
\end{equation}
The second-order response emphasizes changes in the local feature trajectory rather than constant displacement. For a locally linear trajectory \(\hat{x}_{\ell,i}=a_i+\ell b_i\),
\begin{equation}
\hat{x}_{t,i}
-2\hat{x}_{t-1,i}
+\hat{x}_{t-2,i}
=0,
\label{eq:linear_drift}
\end{equation}
so approximately linear feature drift is attenuated. This property does not assume that all camera motion is removed; rather, it reduces persistent linear components and retains deviations associated with changes in the observed motion pattern.

We reshape \(d_t\) into a spatial grid \(M_t\) and apply a fixed spatiotemporal refinement operator consisting of short-term temporal accumulation, morphological closing, and Gaussian smoothing:
\begin{equation}
\tilde{M}_t=\Phi_{\mathrm{st}}(M_t).
\end{equation}

The operator first aggregates recent responses to reduce frame-wise fluctuations. It then connects nearby active regions and smooths their boundaries. These deterministic operations preserve the resolution of \(M_t\), so the motion evidence remains aligned with the token coordinates. Flattening and min--max normalization produce \(\mathcal{S}_{\mathrm{temp},t}\in[0,1]^N\). We use the same fixed refinement across tasks and backbones, without updating any policy parameter.

\subsubsection{Semantic--Motion Agreement and Coverage Modes}

We use the spatial agreement between semantic and motion evidence to select the coverage mode. We first convert the continuous scores into binary masks using adaptive thresholds:

\begin{equation}
\mathcal{B}_{\mathrm{sem},t}
=
\mathbb{I}\!\left(
\mathcal{S}_{\mathrm{sem},t}>
\mu_{\mathrm{sem},t}+\sigma_{\mathrm{sem},t}
\right).
\label{eq:semantic_mask}
\end{equation}
The motion support is obtained analogously:
\begin{equation}
\mathcal{B}_{\mathrm{temp},t}
=
\mathbb{I}\!\left(
\mathcal{S}_{\mathrm{temp},t}>
\mu_{\mathrm{temp},t}+\sigma_{\mathrm{temp},t}
\right).
\label{eq:motion_mask}
\end{equation}
Here, \((\mu_{\mathrm{sem},t},\sigma_{\mathrm{sem},t})\) and \((\mu_{\mathrm{temp},t},\sigma_{\mathrm{temp},t})\) are the spatial mean and standard deviation of the semantic and motion scores, respectively, and \(\mathbb{I}(\cdot)\) is the indicator function. The same one-standard-deviation rule is applied to both maps, yielding score-adaptive masks without introducing separate task-specific threshold coefficients. Their spatial agreement is
\begin{equation}
\operatorname{IoU}_t =
\frac{
\left|
\mathcal{B}_{\mathrm{sem},t}
\cap
\mathcal{B}_{\mathrm{temp},t}
\right|
}{
\left|
\mathcal{B}_{\mathrm{sem},t}
\cup
\mathcal{B}_{\mathrm{temp},t}
\right|
}.
\end{equation}

If the union is empty, we set \(\operatorname{IoU}_t=0\). Let \(\theta_{iou}\) denote the switching threshold. The resulting coverage mode is

\begin{equation}
m_t=
\begin{cases}
\mathrm{Conservative}, & \operatorname{IoU}_t\le\theta_{iou},\\
\mathrm{Aggressive}, & \operatorname{IoU}_t>\theta_{iou}.
\end{cases}
\label{eq:interaction_mode}
\end{equation}
The IoU controls only the coverage mode; it is not used to rank tokens. A low value means that instruction-conditioned support and observed motion have not formed a stable common region. It does not mean that the target is absent. We therefore assign the empty-union case to Conservative Mode. A high value indicates more localized agreement and allows more concentrated coverage, but it is not treated as evidence of physical contact. The mode switch changes only the spatial coverage; the continuous relevance defined below determines the token ordering.

\paragraph{Conservative Mode (Exploration Phase, $\operatorname{IoU}_t \le \theta_{iou}$)}
When semantic intent and observed motion are weakly aligned, a token is treated as background only if neither cue provides high-response support:
\begin{equation}
\mathcal{B}_{\mathrm{bg},t}
=
\neg\mathcal{B}_{\mathrm{sem},t}
\land
\neg\mathcal{B}_{\mathrm{temp},t}.
\label{eq:background_mask}
\end{equation}
The logical AND excludes only regions unsupported by both cues. Equivalently, a region remains active when either semantic intent or observed motion is sufficiently strong. This favors recall while the interaction state is uncertain and avoids requiring premature agreement between the two sources.

\paragraph{Aggressive Mode (Aligned Phase, $\operatorname{IoU}_t > \theta_{iou}$)}
When semantic intent and observed motion are strongly aligned, we retain the core semantic region together with the motion region. Let \(\mathbf{u}_i\) denote the spatial coordinate of token \(i\), \(i_t^*=\arg\max_i\mathcal{S}_{\mathrm{sem},t,i}\), and \(\mathbf{c}_t^*=\mathbf{u}_{i_t^*}\). The core semantic mask is
\(\mathcal{B}_{\mathrm{sem},t}^{\mathrm{core}}
=
\mathcal{B}_{\mathrm{sem},t}
\land
\mathbb{I}(\|\mathbf{u}_i-\mathbf{c}_t^*\|_2\le r)\),
where \(r\) is a fixed spatial radius shared across all evaluations.
The semantic core reduces spatially dispersed instruction responses, while its union with the motion region preserves areas that are currently changing. Geometry is not used in this mode decision and therefore cannot change whether the scheduler uses Conservative or Aggressive coverage.

The two modes produce the provisional active set
\begin{equation}
\small
\mathcal{C}_t =
\begin{cases}
\left\{
x_{t,i}\in X_t^{\mathrm{vis}}
\mid
\mathcal{B}_{\mathrm{bg},t,i}=0
\right\},
& \operatorname{IoU}_t\le\theta_{iou},\\[1mm]
\left\{
x_{t,i}\in X_t^{\mathrm{vis}}
\mid
\mathcal{B}_{\mathrm{sem},t,i}^{\mathrm{core}}
\lor
\mathcal{B}_{\mathrm{temp},t,i}
\right\},
& \operatorname{IoU}_t>\theta_{iou}.
\end{cases}
\end{equation}
The two definitions provide different levels of spatial coverage. Conservative Mode takes the complement of the joint-background region and therefore preserves evidence supported by either cue. Aggressive Mode retains the semantic core and motion support once their agreement becomes sufficiently localized. Geometry is absent from \(\mathcal C_t\) in both cases. Thus, \(\mathcal C_t\) describes only the semantic--motion coverage associated with \(m_t\). It determines the per-frame budget but is neither the final retained sequence nor a hard candidate pool.

\subsubsection{Continuous Interaction Relevance}
The binary supports above are appropriate for measuring spatial agreement, but thresholding discards relative score magnitudes. We therefore retain the continuous semantic and motion responses through the base interaction relevance
\begin{equation}
A_{t,i}
=
\frac{
\mathcal{S}_{\mathrm{sem},t,i}
+
\mathcal{S}_{\mathrm{temp},t,i}
}{2}.
\label{eq:base_relevance}
\end{equation}
By construction, \(A_{t,i}\in[0,1]\). This parameter-free average is used for token ranking, not for selecting the coverage mode or \(K_t\). It provides a dense ordering over all \(N\) existing tokens, including those outside \(\mathcal C_t\). Semantic evidence favors instruction-compatible regions, while motion evidence can retain dynamically relevant regions with weaker semantic responses. In the final stage, geometry adjusts this ordering only through a bounded residual correction. Semantic and motion evidence therefore support both decisions in IAprune: their spatial agreement guides the token count, and their continuous responses determine which tokens compete for that count.

IAprune uses the binary masks only to determine the per-frame budget; they do not directly remove tokens. Specifically, \(m_t\) changes the available token count through \(|\mathcal C_t|\), while \(A_{t,i}\) ranks all \(N\) existing tokens, including those outside \(\mathcal C_t\). Thus, a token excluded from the provisional active set can still be retained if it receives a high continuous relevance score. This separation allows semantic--motion agreement to guide how many tokens are kept without turning the binary masks into a hard candidate pool.

\subsection{Target-Aware Dynamic Token Budget}
Let \(r_{\mathrm{tar}}\in[1/N,1]\) denote the target retention ratio, corresponding to a nominal workload of \(Nr_{\mathrm{tar}}\) tokens per frame. The cardinality \(|\mathcal C_t|\) serves as the interaction-conditioned token demand: Conservative coverage yields a larger value, while localized Aggressive coverage yields a smaller one.

We track the cumulative deviation from the nominal workload by
\begin{equation}
D_{t-1}
=
\sum_{s=1}^{t-1}(K_s-Nr_{\mathrm{tar}}).
\label{eq:budget_account}
\end{equation}
The account starts from \(D_0=0\); its sign indicates cumulative overuse or underuse. To accommodate multi-frame interaction phases, we set its radius to \(B\sqrt t\), where \(B=\min(Nr_{\mathrm{tar}}-1,N-Nr_{\mathrm{tar}})\). The actual budget is the feasible count closest to the current demand:
\begin{equation}
\resizebox{0.85\linewidth}{!}{$
K_t
=
\underset{k\in\{1,\ldots,N\}}{\operatorname{arg\,min}}
\;(k-|\mathcal C_t|)^2
\quad
\mathrm{s.t.}\quad
|D_{t-1}+k-Nr_{\mathrm{tar}}|\le B\sqrt{t}.
$}
\label{eq:dynamic_budget}
\end{equation}
This causal projection preserves the coverage-derived demand within the account envelope and selects the nearest feasible count at its boundary.

For the realized prefix retention \(\bar r_t=(tN)^{-1}\sum_{s=1}^{t}K_s\), the account guarantees
\begin{equation}
\left|\bar r_t-r_{\mathrm{tar}}\right|
\le
\frac{B}{N\sqrt{t}}.
\label{eq:retention_bound}
\end{equation}
Thus, per-frame budgets follow interaction coverage, while their average approaches the target operating point at rate \(\mathcal O(t^{-1/2})\).

\subsection{Budget-Preserving Structural Reallocation}
The global semantic--motion pathway determines both the dynamic budget and the base token ordering, but it can underestimate thin boundaries or texture-poor structures. The wrist view provides a closer observation of these structures. We convert \(I_t^w\) to grayscale and compute its Sobel magnitude
\begin{equation}
\resizebox{0.85\linewidth}{!}{$
\mathcal{E}_t^w(u,v)
=
\sqrt{
\left(I_{t,\mathrm{gray}}^w*K_x\right)^2(u,v)
+
\left(I_{t,\mathrm{gray}}^w*K_y\right)^2(u,v)
},
$}
\end{equation}
where \(*\) denotes 2D convolution and \(K_x,K_y\in\mathbb{R}^{3\times3}\) are the directional Sobel kernels. We then project this response onto the global image plane and aggregate it over the global token grid:
\begin{equation}
\widetilde{E}_t^{\,w\rightarrow g}
=
\operatorname{Norm}\!\left(
\operatorname{Pool}\!\left(
\mathcal{W}_{w\rightarrow g}(\mathcal{E}_t^w)
\right)\right)
\in[0,1]^N.
\label{eq:cross_view_geometry}
\end{equation}
Here, \(\mathcal{W}_{w\rightarrow g}\) is the deterministic depth- and calibration-based reprojection from the wrist camera to the global image plane; \(\operatorname{Pool}\) aggregates the projected response within each global token region, and \(\operatorname{Norm}\) applies min--max normalization. We use the resulting token-level response \(\widetilde{E}_{t,i}^{\,w\rightarrow g}\) to correct the residual confidence not covered by semantic--motion relevance. The final retained sequence is
\begin{equation}
\resizebox{0.85\linewidth}{!}{$
\hat{X}_t
=
\operatorname{TopK}_{K_t}
\left(
\left\{
\left(
x_{t,i},
A_{t,i}+(1-A_{t,i})\widetilde{E}_{t,i}^{\,w\rightarrow g}
\right)
\right\}_{i=1}^{N}
\right).
\label{eq:final_selection}
$}
\end{equation}
The corrected priority remains in \([0,1]\), is monotonic in both \(A_{t,i}\) and \(\widetilde E_{t,i}^{\,w\rightarrow g}\), and introduces no additional fusion coefficient. When \(A_{t,i}\) is high, global semantic and motion evidence dominate the ranking. When its support is limited but the projected wrist observation contains a strong structural boundary, the residual term can promote the corresponding global token. Projected geometry therefore changes which \(K_t\) tokens are retained without changing how many are retained. The selected visual sequence \(\hat{X}_t\) is concatenated with the text tokens and passed to the unchanged embodied manipulation policy for action generation.
% \begin{figure*}[!t]
%     \centering
%     \includegraphics[width=\textwidth]{Benchmark2.pdf}
    
%     \caption{Overview of the evaluation benchmarks and tasks. We evaluate on simulated benchmarks including Libero, VLABench, and CALVIN ABC-D, as well as real-world tasks covering simple, long-horizon, and dual-arm manipulation.}
%     \label{fig:benchmark}
% \end{figure*}
\section{Experiments}

\subsection{Experimental Setup}
We evaluate IAprune in simulation and on a physical robot. In simulation, we apply this training-free method to four open-source embodied manipulation policies: OpenVLA-OFT~\cite{kim2025fine}, DreamVLA~\cite{zhang2025dreamvla}, $\pi_0$~\cite{black2024pi_0}, and $\pi_{0.5}$~\cite{intelligence2025pi05}. We test these policies on LIBERO~\cite{liu2023libero}, VLABench~\cite{zhang2025vlabench}, and CALVIN~\cite{mees2022calvin}. We also evaluate IAprune on a custom-built dual-arm robot platform. All experiments run on Linux with a single NVIDIA A100 GPU.

We use three nominal retention settings: 30\%, 50\%, and 70\%. For IAprune, each setting specifies the target retention ratio \(r_{\mathrm{tar}}\), or a nominal workload of \(Nr_{\mathrm{tar}}\) visual tokens per frame. It does not fix the token count of every frame. Given \(r_{\mathrm{tar}}\), the bounded online scheduler projects the coverage-derived demand \(|\mathcal C_t|\) onto the feasible set defined by the cumulative budget account in Eq.~\eqref{eq:dynamic_budget}. This process requires no target-specific coefficient or calibration set. The resulting budget \(K_t\) can vary with the current semantic--motion coverage, while Eq.~\eqref{eq:retention_bound} keeps its average up to each frame close to \(r_{\mathrm{tar}}\). We compute the observed average retention from the realized budgets over all evaluated frames and episodes. For each baseline, the same nominal percentage follows the budget definition in its original implementation.

\subsection{Evaluation Benchmarks}
\subsubsection{\textbf{Simulation Benchmarks}}
\begin{figure*}[!t]
    \centering
    \includegraphics[width=0.9\textwidth]{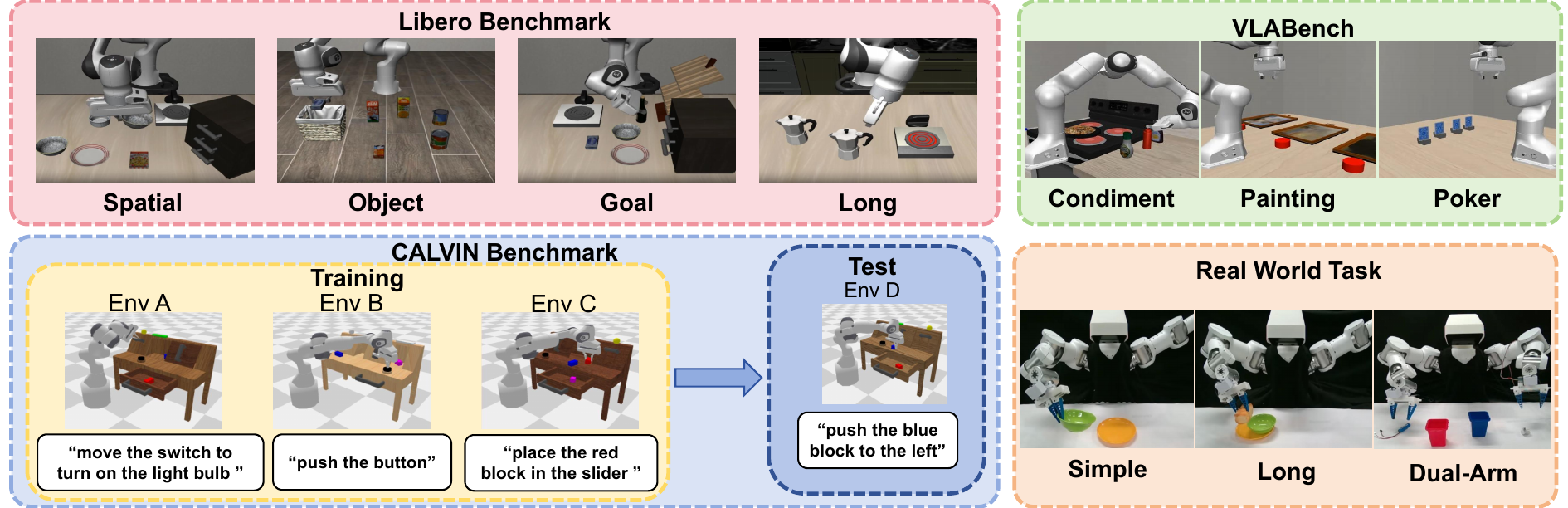}
    
       \caption{Overview of the evaluation benchmarks and tasks. The simulation evaluation includes LIBERO, VLABench, and CALVIN ABC-D. The real-world evaluation covers simple, long-horizon, and dual-arm manipulation.}

    \label{fig:benchmark}
\end{figure*}

As shown in Fig.~\ref{fig:benchmark}, we use three simulation benchmarks that cover different aspects of embodied manipulation. LIBERO~\cite{liu2023libero} contains spatial, object, goal, and long-horizon task suites. CALVIN~\cite{mees2022calvin} evaluates long-horizon, language-conditioned manipulation through sequences of consecutive tasks. VLABench~\cite{zhang2025vlabench} includes complex manipulation tasks that require long-horizon reasoning. For VLABench, the main paper reports results on three representative task subsets: add\_condiment, select\_painting, and select\_poker.

\begin{figure}[!t]
    \centering
    \includegraphics[width=0.9\columnwidth]{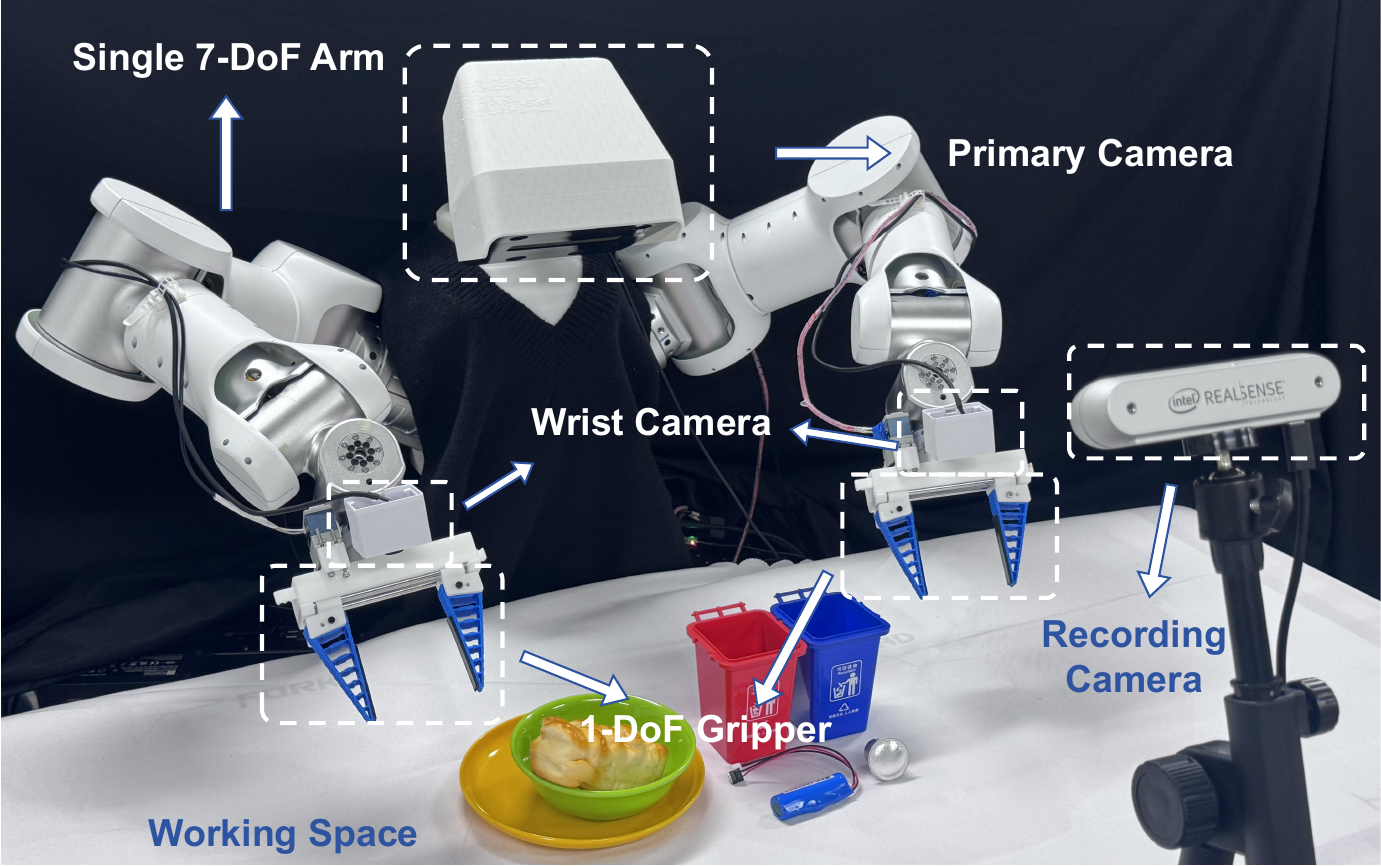}
    \caption{Physical platform for real-world evaluation. The head-mounted primary camera and wrist cameras provide complementary observations to the embodied manipulation policy, while the external RealSense camera is used only to record task execution. The shared workspace supports both single-arm manipulation and coordinated dual-arm operation.}
    \label{fig:robot_setup}
\end{figure}
\subsubsection{\textbf{Real-World Experiments}}

To evaluate the generalizability and real-time control capability of IAprune, we deploy it on a physical robotic platform and conduct experiments under both single-arm and dual-arm settings, as shown in Fig.~\ref{fig:robot_setup}. 
The deployment hardware, camera configuration, and inference protocol are summarized in Table~\ref{tab:real_robot_setup}. 
We design three manipulation tasks with increasing complexity: (1)~\textit{Simple}, where the robot picks a bowl and places it onto a plate; (2)~\textit{Long}, a multi-step sequence that first places a bowl onto a plate and then places bread into the bowl; and (3)~\textit{Dual-arm}, where two arms coordinate to sort hazardous and non-hazardous waste into the corresponding bins. 
We compare IAprune with the unpruned $\pi_{0.5}$ backbone and representative pruning baselines under the same deployment protocol.

\begin{table}[!t]
\centering
\scriptsize
\renewcommand{\arraystretch}{0.82}
\setlength{\tabcolsep}{2.0pt}
\caption{\textbf{Real-Robot Deployment Details.} Hardware and inference configuration used for physical experiments.}
\label{tab:real_robot_setup}
\begin{tabular}{@{}p{0.25\columnwidth}p{0.70\columnwidth}@{}}
\toprule[0.9pt]
\textbf{Robot Platform} & Self-developed dual-arm system; 2$\times$7-DoF arms + 1-DoF parallel-jaw gripper; fixed base \\
\midrule[0.5pt]
\textbf{Global View} & ZED 2i binocular camera mounted on the head; RGB stream from the left lens only \\
\textbf{Wrist View} & Wide-angle RGB wrist camera per arm; single-arm uses head + active wrist; dual-arm uses head + both wrists \\
\textbf{Data Spec.} & All observations standardized to 640$\times$480; Intel RealSense D455 used only for external recording \\
\midrule[0.5pt]
\textbf{OS / Hardware} & Ubuntu 22.04 workstation with a single NVIDIA A100 GPU \\
\textbf{Backbone} & $\pi_{0.5}$ \\
\textbf{Inference} & Action chunk size 50; first 20 actions of each predicted chunk are executed \\
\bottomrule[0.9pt]
\end{tabular}
\end{table}

\subsection{Main Results on Simulation Benchmarks}
\subsubsection{\textbf{Cross-Benchmark Comparison}}

Table~\ref{tab:comprehensive_comparison} compares IAprune against baselines across three nominal retention settings. Evaluation spans three embodied manipulation policies and three benchmarks focusing on generalization, long-horizon actions, and complex reasoning. While all methods degrade as the nominal retention decreases, IAprune demonstrates superior robustness. At the nominal 70\% setting, IAprune surpasses the unpruned baseline on multiple metrics, achieving a 4.45 average sequence length on CALVIN and 46.3\% success rate on VLABench. This result is consistent with the dynamic scheduler preserving broader context during uncertain phases while reducing the average visual-token workload.
\par
Performance divergence becomes critical at the nominal 30\% setting on the challenging VLABench. VLABench tasks require precise geometric perception and fine manipulation control. Under extreme compression, the reported success rates of the perception-first baselines decrease sharply, with DivPrune reaching 6.7\%. Although VLA-Cache accounts for temporal continuity by caching historical features, its success rate decreases from 40.7\% at the nominal 70\% setting to 8.0\% at the nominal 30\% setting. In contrast, IAprune maintains a 33.3\% success rate at the same nominal operating point. These results show that IAprune is more robust under the evaluated aggressive-compression setting and are consistent with interaction-aligned scheduling and budget-preserving structural reallocation retaining manipulation-relevant evidence.

\newcommand{\bestbase}[1]{\textbf{#1}}
\begin{table*}[!t]
\centering
\scriptsize
\renewcommand{\arraystretch}{1.02}
\setlength{\tabcolsep}{1.4pt}
\caption{\textbf{Comprehensive Performance Comparison.}
Success rates are reported on LIBERO and VLABench, and average sequence length is reported on CALVIN at nominal retention labels of 30\%, 50\%, and 70\%. For IAprune, these labels denote target average retention ratios. Baseline labels follow their original budget definitions and may not correspond to identical realized token counts.}
\label{tab:comprehensive_comparison}
\resizebox{\textwidth}{!}{%
\begin{tabular}{l|cccc|cccccc|cccc|cccc}
\hline
\multirow{3}{*}{\textbf{Method}} 
& \multicolumn{10}{c|}{\textbf{DreamVLA}} 
& \multicolumn{8}{c}{\textbf{OpenPi} $(\pi_0,\pi_{0.5})$} \\
\cline{2-19}
& \multicolumn{4}{c|}{\textbf{LIBERO Success (\%)}} 
& \multicolumn{6}{c|}{\textbf{CALVIN Seq. Length} $\uparrow$}
& \multicolumn{4}{c|}{\textbf{LIBERO Success (\%)}}
& \multicolumn{4}{c}{\textbf{VLABench Success (\%)}} \\
\cline{2-19}
& Spa. & Obj. & Goal & Long 
& 1 & 2 & 3 & 4 & 5 & Avg.
& Spa. & Obj. & Goal & Long 
& Con. & Paint. & Poker. & Avg. \\
\hline

\multicolumn{19}{c}{\cellcolor[gray]{0.9}\textbf{Vanilla Baseline: 100\% Token Retention}} \\
\hline
Vanilla 
& 84.5 & 91.5 & 89.5 & 89.5 
& 97.8 & 94.5 & 89.5 & 83.8 & 77.5 & 4.43 
& 94.2 & 98.2 & 94.6 & 82.8 
& 56.0 & 30.0 & 52.0 & 46.0 \\
\hline

\multicolumn{19}{c}{\cellcolor[gray]{0.9}\textbf{Retention: 70\%}} \\
\hline
FastV 
& 86.5 & 86.5 & 84.0 & 87.5 
& 97.6 & 94.5 & 89.0 & 82.8 & 76.8 & 4.40 
& 93.5 & 96.1 & 92.2 & 80.5 
& 44.0 & 28.0 & 32.0 & 34.6 \\
SparseVLM 
& 86.5 & 85.0 & 91.5 & 86.5 
& 97.4 & 93.8 & 88.6 & 83.0 & 76.8 & 4.40 
& 93.8 & 94.8 & 92.5 & 80.1 
& 50.0 & 32.0 & 26.0 & 36.0 \\
DivPrune 
& 85.5 & 90.5 & 88.0 & 85.5 
& 98.0 & 94.3 & 88.9 & 83.4 & \bestbase{78.3} & 4.42 
& 91.8 & 92.5 & 91.4 & 79.2 
& 6.0 & 22.0 & 2.0 & 10.0 \\
VLA-Cache 
& 88.0 & 89.5 & 89.5 & 87.5 
& 98.2 & 94.8 & 89.2 & 82.5 & 77.1 & 4.42 
& 94.5 & 97.8 & 94.5 & 82.6 
& 42.0 & 32.0 & \bestbase{48.0} & 40.7 \\
\textbf{IAprune} 
& \bestbase{91.5} & \bestbase{93.5} & \bestbase{92.5} & \bestbase{90.5} 
& \bestbase{98.4} & \bestbase{95.1} & \bestbase{89.7} & \bestbase{83.7} & 78.0 & \bestbase{4.45} 
& \bestbase{94.8} & \bestbase{98.5} & \bestbase{95.2} & \bestbase{84.2} 
& \textbf{55.0} & \bestbase{36.0} & \bestbase{48.0} & \bestbase{46.3} \\
\hline

\multicolumn{19}{c}{\cellcolor[gray]{0.9}\textbf{Retention: 50\%}} \\
\hline
FastV 
& 84.5 & 81.5 & 84.0 & 86.5 
& 94.2 & 92.8 & 84.5 & 79.7 & 70.1 & 4.21 
& 92.4 & 95.3 & 91.0 & 79.4 
& 16.0 & 22.0 & 2.0 & 13.3 \\
SparseVLM 
& 84.5 & 83.5 & \bestbase{90.5} & 86.5 
& 97.9 & 94.2 & 89.1 & 82.9 & 76.1 & 4.40 
& 92.0 & 93.8 & 90.7 & 77.6 
& 14.0 & 30.0 & 2.0 & 15.3 \\
DivPrune 
& 84.5 & \bestbase{92.0} & 84.5 & 85.0 
& 98.3 & 93.4 & 84.9 & 80.3 & 74.3 & 4.32 
& 88.8 & 92.0 & 89.8 & 75.3 
& 4.0 & 22.0 & 2.0 & 9.3 \\
VLA-Cache 
& 87.5 & 88.0 & 86.0 & 87.5 
& 98.1 & 92.5 & 89.0 & 82.1 & 76.8 & 4.39 
& 94.1 & 95.9 & \bestbase{94.8} & 82.0 
& 16.8 & 30.0 & 11.3 & 19.3 \\
\textbf{IAprune} 
& \bestbase{90.5} & \bestbase{92.0} & \bestbase{90.5} & \bestbase{89.7} 
& \bestbase{98.4} & \bestbase{94.3} & \bestbase{89.4} & \bestbase{83.5} & \bestbase{77.9} & \bestbase{4.44} 
& \bestbase{94.5} & \bestbase{98.3} & 94.5 & \bestbase{83.4} 
& \bestbase{48.9} & \bestbase{32.0} & \bestbase{42.0} & \bestbase{41.0} \\
\hline

\multicolumn{19}{c}{\cellcolor[gray]{0.9}\textbf{Retention: 30\%}} \\
\hline
FastV 
& 79.0 & 56.5 & 73.0 & 81.5 
& 89.8 & 85.2 & 78.2 & 71.5 & 63.5 & 3.88 
& 86.8 & 80.7 & 86.2 & 73.9 
& 0.0 & 22.0 & 0.0 & 7.3 \\
SparseVLM 
& 76.5 & 43.0 & 79.0 & 80.5 
& 96.3 & 90.9 & 83.5 & 74.4 & 66.3 & 4.11 
& 87.1 & 86.2 & 82.5 & 71.9 
& 18.0 & 26.0 & 2.0 & 15.3 \\
DivPrune 
& 71.0 & 87.5 & 81.5 & 80.5 
& 96.8 & 90.8 & 85.0 & 76.8 & 69.8 & 4.19 
& 82.1 & 72.2 & 79.4 & 66.0 
& 4.0 & 16.0 & 0.0 & 6.7 \\
VLA-Cache 
& 81.5 & 81.5 & 79.2 & 79.5 
& 97.1 & 90.2 & 82.6 & 78.5 & 68.6 & 4.17 
& 86.5 & 87.2 & 85.6 & 72.6 
& 0.0 & 24.0 & 0.0 & 8.0 \\
\textbf{IAprune} 
& \bestbase{87.5} & \bestbase{89.5} & \bestbase{89.2} & \bestbase{89.1} 
& \bestbase{97.8} & \bestbase{94.7} & \bestbase{89.6} & \bestbase{83.5} & \bestbase{76.4} & \bestbase{4.42} 
& \bestbase{93.2} & \bestbase{97.6} & \bestbase{94.2} & \bestbase{82.6} 
& \bestbase{46.0} & \bestbase{32.0} & \bestbase{22.0} & \bestbase{33.3} \\
\hline
\end{tabular}}
\end{table*}

\subsubsection{\textbf{Detailed Comparison on LIBERO}}

Table~\ref{tab:libero_comprehensive} presents the performance of various embodied manipulation policies and visual-token pruning methods on the LIBERO benchmark. The unpruned OpenVLA-OFT policy establishes a strong baseline with a 97.1\% average success rate. At the target average retention \(r_{\mathrm{tar}}=70\%\), IAprune achieves a 97.8\% average success rate and a $1.25\times$ inference speedup. EfficientVLA delivers a $1.54\times$ speedup with an 88.9\% average success rate. At the more aggressive target \(r_{\mathrm{tar}}=30\%\), IAprune achieves the same reported average success rate as the unpruned policy (97.1\%) with a $1.54\times$ speedup. These results show that the target average retention ratio provides controllable operating points along the accuracy--efficiency trade-off.

\definecolor{SectionGray}{gray}{0.975}
\definecolor{OursGray}{gray}{0.985}
\newcommand{\nada}{-}
\newcommand{\best}[1]{\textbf{#1}}

\begin{table}[!t]
\centering
\scriptsize
\renewcommand{\arraystretch}{1.01}
\setlength{\tabcolsep}{1.5pt}

\caption{\textbf{LIBERO results for embodied manipulation policies and visual-token pruning methods.}
Representative policies are included for context. All pruning methods use the OFT 7B backbone, and speedup is reported relative to the unpruned OpenVLA-OFT policy.}

\label{tab:libero_comprehensive}

\resizebox{\columnwidth}{!}{%
\begin{tabular}{@{}l l ccccc c@{}}
\hline
\multirow{2}{*}{\textbf{Method}} &
\multirow{2}{*}{\textbf{CKPT}} &
\multicolumn{5}{c}{\textbf{Success Rate (\%)}} &
\multirow{2}{*}{\textbf{Speedup}} \\
\cline{3-7}
& & Spa. & Obj. & Goal & Long & \textbf{Avg.} & \\
\hline

\rowcolor{SectionGray}
\multicolumn{8}{@{}c@{}}{\textbf{Embodied Manipulation Policies}} \\
\hline
OpenVLA~\cite{kim2024openvla} 
& OpenVLA (7B) & 84.7 & 88.4 & 79.2 & 53.7 & 76.5 & \nada \\
WorldVLA~\cite{cen2025worldvla} 
& Cham. (7B) & 87.6 & 96.2 & 83.4 & 60.0 & 81.8 & \nada \\ 
NORA~\cite{hung2025nora} 
& Qwen-VL (3B) & 85.6 & 87.8 & 77.0 & 45.0 & 73.9 & \nada \\
SmolVLA~\cite{shukor2025smolvla} 
& SmolVLM (2.2B) & 93.0 & 94.0 & 91.0 & 77.0 & 88.8 & \nada \\ 
CogACT~\cite{li2024cogact} 
& CogVLM (7B) & 97.2 & 98.0 & 90.2 & 88.8 & 93.6 & \nada \\
OpenVLA-OFT~\cite{kim2025fine} 
& OFT (7B) & 98.6 & 98.2 & 96.6 & 94.8 & 97.1 & 1.00$\times$ \\
\hline

\rowcolor{SectionGray}
\multicolumn{8}{@{}c@{}}{\textbf{Pruning Methods}} \\
\hline
VLA-ADP~\cite{pei2025action} (Ret. 70\%) 
& OFT (7B) & 99.0 & 98.2 & 96.8 & 91.2 & 96.3 & 1.13$\times$ \\
VLA-ADP~\cite{pei2025action} (Ret. 50\%) 
& OFT (7B) & \best{99.4} & 98.0 & 96.4 & 91.2 & 96.3 & 1.23$\times$ \\
VLA-ADP~\cite{pei2025action} (Ret. 30\%) 
& OFT (7B) & 97.6 & 98.4 & 97.4 & 84.2 & 94.4 & 1.35$\times$ \\
FastV~\cite{chen2024image} 
& OFT (7B) & 96.8 & 81.0 & 96.4 & 73.0 & 86.8 & 1.24$\times$ \\
VLA-Cache~\cite{xu2025vla} 
& OFT (7B) & 98.3 & 97.5 & 98.3 & 95.4 & 97.4 & 1.30$\times$ \\
SpecPrune-VLA~\cite{wang2025specprune} 
& OFT (7B) & 98.2 & 96.3 & 97.7 & 94.0 & 96.6 & 1.46$\times$ \\
TeamVLA~\cite{ye2025token} 
& OFT (7B) & 99.2 & 96.5 & 97.0 & 93.8 & 96.6 & 1.51$\times$ \\
EfficientVLA~\cite{yang2025efficientvla} 
& OFT (7B) & 96.5 & 91.1 & 96.0 & 72.1 & 88.9 & \best{1.54$\times$} \\
\rowcolor{OursGray}
\textbf{IAprune (Ret. 70\%)} 
& OFT (7B) & 97.6 & \best{99.6} & \best{98.4} & \best{95.6} & \best{97.8} & 1.25$\times$ \\
\rowcolor{OursGray}
\textbf{IAprune (Ret. 50\%)} 
& OFT (7B) & 97.3 & 99.1 & 98.2 & 95.2 & 97.5 & 1.37$\times$ \\
\rowcolor{OursGray}
\textbf{IAprune (Ret. 30\%)} 
& OFT (7B) & 96.6 & 98.8 & 98.0 & 94.8 & 97.1 & \best{1.54$\times$} \\
\hline
\end{tabular}}
\vspace{-1mm}
\end{table}

% \begin{figure}[!t]
%     \centering
%     \includegraphics[width=\textwidth]{"visual compare2.pdf"}
%     \caption{\textbf{Qualitative comparison on LIBERO under extreme token pruning.} At a target average retention of 12.5\%, IAprune first uses Conservative Mode with background-only filtering to preserve the target bowl during the uncertain early phase, then maintains connected tokens around the manipulator and target after interaction alignment. Baseline pruning methods lose target-adjacent tokens or keep scattered redundant patches, leading to weaker preservation of task-critical regions.}
%     \label{fig:libero_visual_compare}
% \end{figure}

\subsection{Analysis of Interaction-Aligned Pruning}
\subsubsection{\textbf{Diagnosing Evidence Misalignment in Token Pruning}}

Reducing the token count is useful only if the retained sequence still covers visual evidence needed for control. We therefore examine three regions that may contain such evidence: the instruction-relevant target, its boundary, and the potential robot--object interaction zone. On VLABench, we measure their coverage using three post-hoc, segmentation-grounded proxies. Target-object token retention (TTR) is the fraction of target-object tokens retained after pruning. Target-boundary token retention (BTR) measures the retained fraction along the target boundary, and contact-zone retention (CZR) measures the retained fraction near the potential robot--target interaction region. These metrics do not represent the causal importance of individual tokens, which is not directly observable. They are used only as coverage proxies. We construct the masks from VLABench task metadata, MuJoCo geometry IDs, robot masks, and per-frame segmentation maps. Neither the policy nor any pruning method uses these masks during inference.

\begin{table}[!t]
\centering
\renewcommand{\arraystretch}{0.98}
\setlength{\tabcolsep}{1.1pt}
\caption{\textbf{Segmentation-grounded token retention on VLABench.}
TTR, BTR, and CZR measure the retained fractions of target-object, target-boundary, and contact-zone tokens. The early segment is the first 30\% of each episode, and the late segment is the remaining 70\%.}

\label{tab:action_critical_retention_seg_early_late}
\begin{tabular*}{\columnwidth}{@{\extracolsep{\fill}}llcccccc@{}}
\toprule[0.9pt]
\multirow{2}{*}{\textbf{Ret.}} & \multirow{2}{*}{\textbf{Method}}
& \multicolumn{3}{c}{\textbf{Early segment}}
& \multicolumn{3}{c}{\textbf{Late segment}} \\
\cmidrule(lr){3-5}\cmidrule(lr){6-8}
& & \textbf{TTR}$\uparrow$ & \textbf{BTR}$\uparrow$ & \textbf{CZR}$\uparrow$
& \textbf{TTR}$\uparrow$ & \textbf{BTR}$\uparrow$ & \textbf{CZR}$\uparrow$ \\
\midrule[0.6pt]
30\% & SparseVLM & 0.016 & 0.049 & 0.100 & 0.077 & 0.083 & 0.125 \\
 & VLA-Cache & 0.267 & 0.271 & 0.312 & 0.304 & 0.395 & 0.356 \\
 & DivPrune & 0.374 & 0.311 & 0.313 & 0.444 & 0.415 & 0.400 \\
 & \textbf{IAprune} & \textbf{0.733} & \textbf{0.672} & \textbf{0.710} & \textbf{0.543} & \textbf{0.507} & \textbf{0.565} \\
\midrule[0.5pt]
50\% & SparseVLM & 0.207 & 0.334 & 0.571 & 0.273 & 0.391 & 0.545 \\
 & VLA-Cache & 0.524 & 0.536 & 0.580 & 0.663 & 0.703 & 0.639 \\
 & DivPrune & 0.563 & 0.517 & 0.516 & 0.614 & 0.594 & 0.576 \\
 & \textbf{IAprune} & \textbf{0.819} & \textbf{0.814} & \textbf{0.810} & \textbf{0.710} & \textbf{0.722} & \textbf{0.758} \\
\midrule[0.5pt]
70\% & SparseVLM & 0.877 & 0.819 & \textbf{0.969} & 0.899 & 0.856 & \textbf{0.963} \\
 & VLA-Cache & 0.782 & 0.778 & 0.811 & 0.847 & 0.863 & 0.817 \\
 & DivPrune & 0.738 & 0.696 & 0.678 & 0.787 & 0.751 & 0.704 \\
 & \textbf{IAprune} & \textbf{0.899} & \textbf{0.906} & 0.901 & \textbf{0.903} & \textbf{0.884} & 0.881 \\
\bottomrule[0.9pt]
\end{tabular*}
\end{table}

\begin{figure*}[!t]
    \centering
    \includegraphics[width=\textwidth]{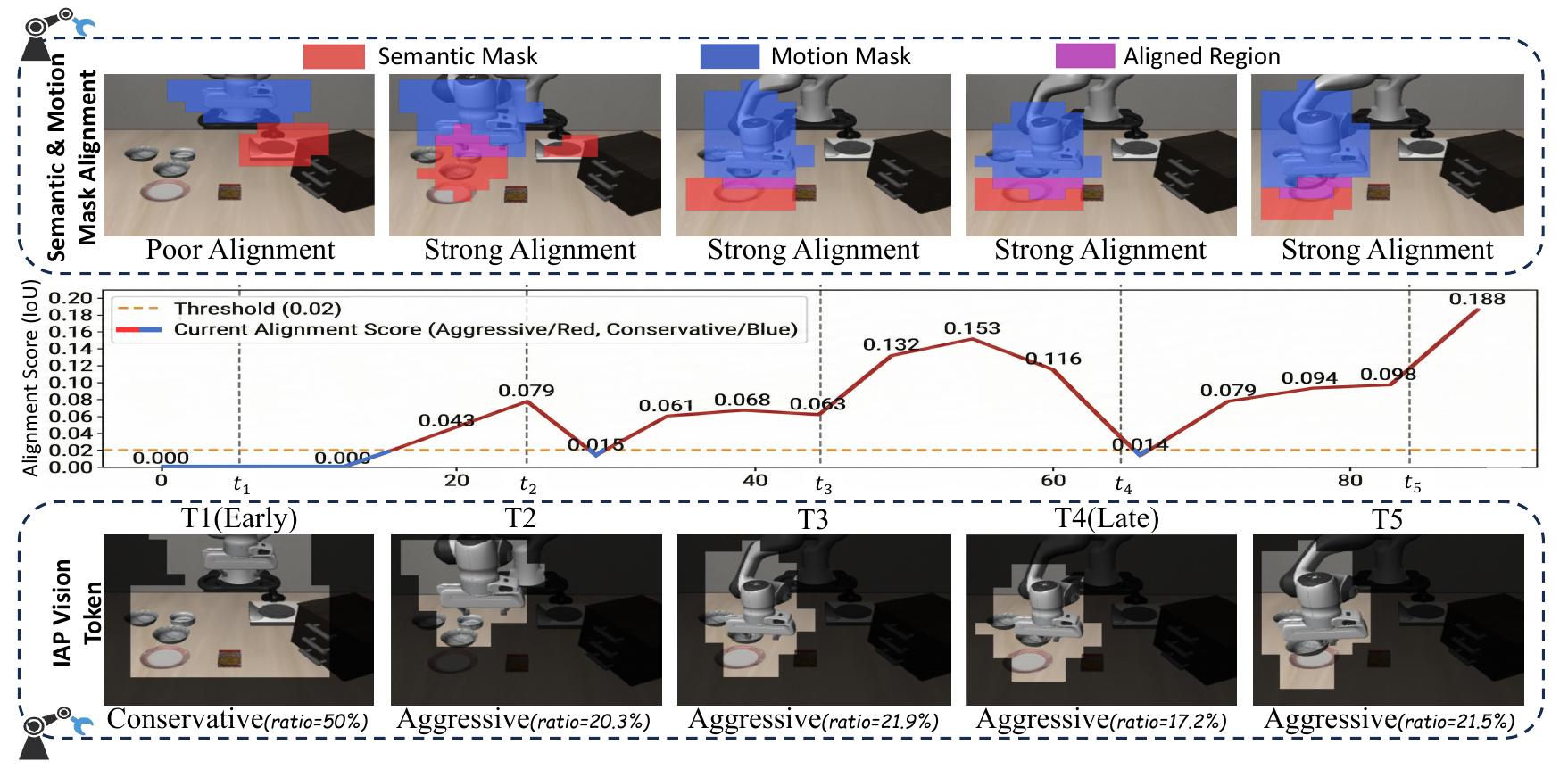}
    
       \caption{\textbf{Visualization of interaction-aligned pruning on LIBERO.} Bottom: visual-token retention changes from Conservative to Aggressive coverage. Middle: the alignment score (IoU) selects the coverage mode. Top: the overlap (purple) between the semantic (red) and observed-motion (blue) masks determines the alignment score.}

    \label{fig:visualization_pruning}
\end{figure*}

Table~\ref{tab:action_critical_retention_seg_early_late} shows how evidence retention changes across rollout segments and token budgets. At 30\%, the strongest baseline reaches early-segment TTR/BTR/CZR values of 0.374/0.311/0.313, while IAprune reaches 0.733/0.672/0.710. The gap remains clear at 50\%, especially for boundary and contact-zone evidence. At 70\%, the gap narrows, and SparseVLM attains the highest CZR. Thus, baseline compression cues can preserve contact-zone evidence when more tokens are available. Overall, the benefit of IAprune is largest in the early segment under a tight token budget. This pattern is consistent with using weak semantic--motion agreement to retain broader coverage rather than switching too early to concentrated coverage.

% \begin{figure}[!t]
% \centering
% \includegraphics[width=\columnwidth]{vlabench_llm_final.pdf}
% \caption{\textbf{Attention analysis of the vision-language backbone on VLABench.} Heatmaps of $\pi_{0.5}$ internal attention across layers 8, 12, and 16 show that the backbone can focus on irrelevant background regions rather than task-critical objects in complex manipulation tasks.}
% \label{fig:attention_fail}
% \end{figure}

% Fig.~\ref{fig:attention_fail} provides a complementary qualitative diagnosis: internal attention can concentrate on background regions rather than the physical target, so pruning rules tied to attention, saliency, redundancy, or diversity may spend the scarce early budget on visually prominent but action-irrelevant tokens. IAprune is designed around this failure mode: it delays aggressive pruning until semantic and motion evidence become spatially aligned, then keeps sparse boundary evidence represented through structural reallocation within the selected budget.

\subsubsection{\textbf{Dynamic Conservative-to-Aggressive Pruning}}

Fig.~\ref{fig:visualization_pruning} shows coverage-mode switching during one complete LIBERO episode. At \(t_1\), the semantic and motion masks do not overlap, so the IoU is zero and remains below the switching threshold. The IoU rule therefore selects Conservative Mode. In this example, Conservative Mode produces a relatively large provisional set \(\mathcal C_t\) and a larger coverage-derived token demand. From \(t_2\) onward, the two masks overlap and the IoU rises above the threshold. The scheduler then selects Aggressive Mode, which narrows the provisional coverage and lowers the token demand. The bounded budget account maps this demand to \(K_t\) while keeping the cumulative deviation within the target-retention bound.

\subsubsection{\textbf{Effect of Budget-Preserving Structural Reallocation}}

\begin{figure}[!t]
\centering
\includegraphics[width=\columnwidth]{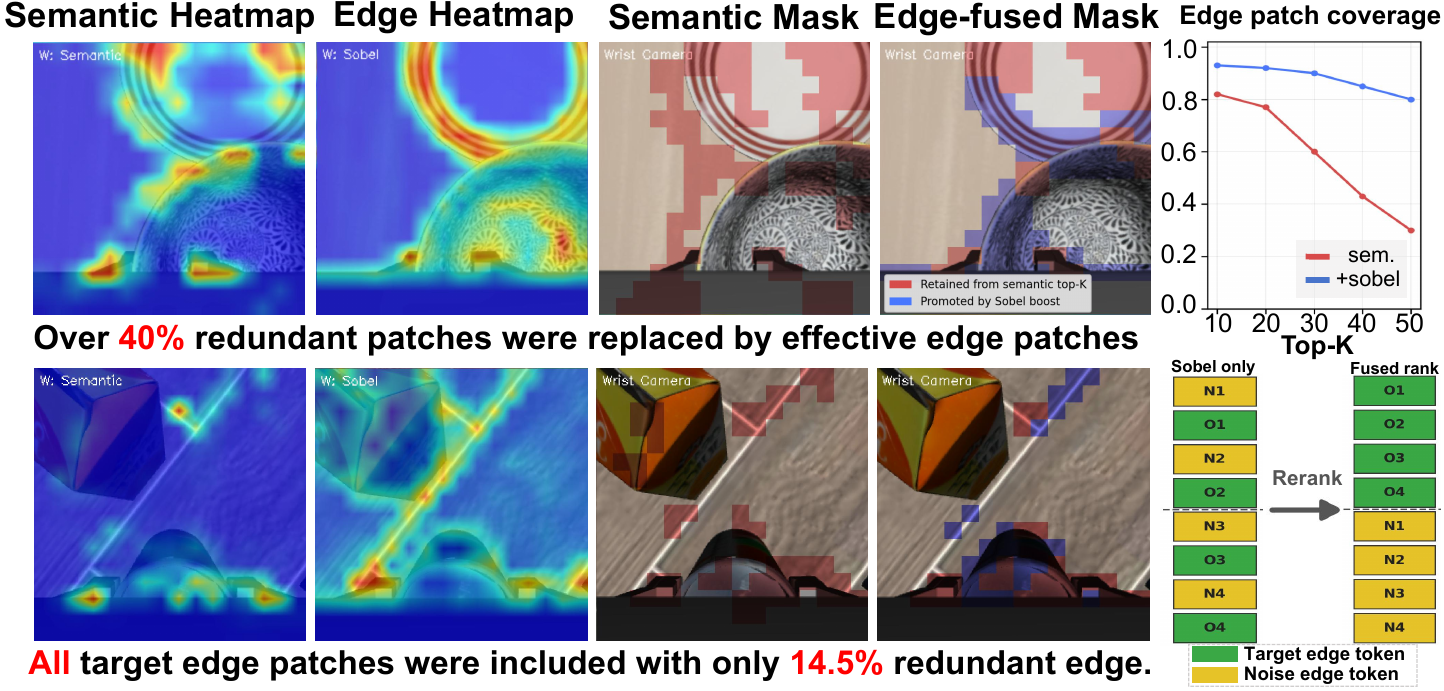}
\caption{\textbf{Effect of budget-preserving structural reallocation.} Under the same per-frame Top-$K_t$ budget, geometry-guided residual correction retains more object-boundary tokens near potential contact regions.}

\label{fig:edge_visualization}
\end{figure}

We next ask whether structural reallocation changes which tokens are retained under a fixed per-frame budget. This controlled comparison differs from the full scheduler, where \(K_t\) varies over time. Fig.~\ref{fig:edge_visualization} compares semantic-only selection with geometry-guided residual correction on representative frames, using \(K_t=77\) in both cases. Semantic-only selection retains most object-body tokens but misses many thin boundaries with weak semantic responses. Structural reallocation redirects some of the same Top-$K_t$ slots toward these geometric cues. As shown in Table~\ref{tab:edge_token_allocation}, the average number of retained object-edge tokens increases from 7.9 to 24.1, while the combined number of semantic-drift, noise-edge, and background tokens decreases from 47.8 to 30.9. The total remains 77 tokens in both cases. Geometry therefore changes which existing tokens are retained without increasing the sequence length.

\begin{table}[!t]
\centering
\scriptsize
\renewcommand{\arraystretch}{0.78}
\setlength{\tabcolsep}{1.2pt}
\caption{\textbf{Budget-preserving structural reallocation.} Geometry-guided residual correction reallocates a fixed budget toward object edges near potential contact regions. Both variants use \(K_t=77\).}

\label{tab:edge_token_allocation}
\resizebox{\columnwidth}{!}{%
\begin{tabular}{@{}lcccc@{}}
\toprule[0.9pt]
\textbf{Token type} & \textbf{Available} & \textbf{Semantic only} & \textbf{+Sobel} & \textbf{Avg. change} \\
\midrule[0.6pt]
Object body            & 22  & 21.3\,(97\%) & 22.0\,(100\%) & $+0.7$ \\
Semantic drift         & 18  & 14.5\,(81\%) & \phantom{0}6.8\,(38\%) & $-7.7$ \\
\textbf{Object edge}   & 26  & \phantom{0}7.9\,(30\%) & \textbf{24.1\,(93\%)} & $\mathbf{+16.2}$ \\
\textbf{Noise edge}    & 38  & \phantom{0}2.6\,(7\%) & \phantom{0}6.4\,(17\%) & $+3.8$ \\
Other/background       & 152 & 30.7\,(20\%) & 17.7\,(12\%) & $-13.0$ \\
\midrule[0.5pt]
\textbf{Top-$K$ total} & \textbf{256\,(100\%)} & \textbf{77.0\,(keep30\%)} & \textbf{77.0\,(keep30\%)} & -- \\
\bottomrule[0.8pt]
\end{tabular}}
\end{table}

\subsection{Ablation and Robustness}
\subsubsection{\textbf{Component Ablation}}

Table~\ref{tab:mechanism_ablation} isolates the main components of IAprune across three backbones and three target retention ratios. For each backbone and target \(r_{\mathrm{tar}}\), all variants are calibrated to the same average retention, so performance differences do not come from retaining more tokens on average. Adding motion or edge evidence alone produces small and sometimes inconsistent changes over the semantic-only variant. Combining both cues without coverage-mode switching also recovers only part of the gain. Full IAprune achieves the highest success rate in every setting. At 30\% retention, it improves over the semantic-only variant by 6.1, 8.0, and 7.4 percentage points on OpenVLA-OFT, DreamVLA, and \(\pi_0\), respectively. These gains narrow to 2.5, 3.3, and 2.2 points at 70\%. The variant without structural reallocation retains most of these gains, suggesting that semantic--motion scheduling is most useful when the token budget is tight. Removing structural reallocation still lowers success by 0.6--1.6 points across the table, showing a smaller but consistent complementary gain.

\begin{table}[!t]
\centering
\renewcommand{\arraystretch}{0.91}
\setlength{\tabcolsep}{0.6pt}
\caption{\textbf{Mechanism ablation across backbones.} Average success rate under matched average retention. Sem. is the semantic-only variant; +Mot. and +Edge add one cue at a time; +M+E combines both cues without coverage-mode switching; and w/o Edge. removes structural reallocation from full IAprune. Parentheses report percentage-point changes from Sem.}

\label{tab:mechanism_ablation}
\begin{tabular*}{\columnwidth}{@{\extracolsep{\fill}}lcccccc@{}}
\toprule[0.9pt]
\textbf{Budget} & \textbf{Sem.} 
& \textbf{+Mot.} 
& \textbf{+Edge} 
& \textbf{+M+E} 
& \textbf{Full IAP} 
& \textbf{w/o Edge} \\
\midrule[0.7pt]

\rowcolor{SectionGray}
\multicolumn{7}{c}{\textbf{OpenVLA-OFT}} \\
30\% & 91.0 & 90.7 (-0.3) & 91.8 (+0.8) & 91.4 (+0.4) & \textbf{97.1 \textcolor{red}{(+6.1)}} & 96.1 (+5.1) \\
50\% & 94.0 & 94.0 (+0.0) & 94.8 (+0.8) & 94.2 (+0.2) & \textbf{97.5 \textcolor{red}{(+3.5)}} & 96.8 (+2.8) \\
70\% & 95.3 & 95.5 (+0.2) & 95.9 (+0.6) & 96.0 (+0.7) & \textbf{97.8 \textcolor{red}{(+2.5)}} & 97.2 (+1.9) \\
\midrule[0.6pt]

\rowcolor{SectionGray}
\multicolumn{7}{c}{\textbf{DreamVLA}} \\
30\% & 80.8 & 80.2 (-0.6) & 82.0 (+1.2) & 81.5 (+0.7) & \textbf{88.8 \textcolor{red}{(+8.0)}} & 87.2 (+6.4) \\
50\% & 86.0 & 85.8 (-0.2) & 87.2 (+1.2) & 86.3 (+0.3) & \textbf{90.7 \textcolor{red}{(+4.7)}} & 89.8 (+3.8) \\
70\% & 88.7 & 88.4 (-0.3) & 89.4 (+0.7) & 89.0 (+0.3) & \textbf{92.0 \textcolor{red}{(+3.3)}} & 91.0 (+2.3) \\
\midrule[0.6pt]

\rowcolor{SectionGray}
\multicolumn{7}{c}{$\boldsymbol{\pi}_0$} \\
30\% & 84.5 & 83.8 (-0.7) & 85.4 (+0.9) & 85.8 (+1.3) & \textbf{91.9 \textcolor{red}{(+7.4)}} & 90.6 (+6.1) \\
50\% & 88.2 & 87.5 (-0.7) & 88.6 (+0.4) & 88.7 (+0.5) & \textbf{92.7 \textcolor{red}{(+4.5)}} & 91.7 (+3.5) \\
70\% & 91.0 & 90.6 (-0.4) & 91.5 (+0.5) & 91.4 (+0.4) & \textbf{93.2 \textcolor{red}{(+2.2)}} & 92.2 (+1.2) \\

\bottomrule[0.9pt]
\end{tabular*}
\end{table}

\subsubsection{\textbf{Budget Tracking and Threshold Sensitivity}}

Table~\ref{tab:robustness_scheduler} first examines how closely the bounded online scheduler follows the target retention ratio. For \(r_{\mathrm{tar}}\in\{30\%,50\%,70\%\}\), the observed averages are 30.9\%, 49.7\%, and 67.7\%. The corresponding deviations are \(+0.9\), \(-0.3\), and \(-2.3\) percentage points. Across the evaluated IoU thresholds, the observed retention ranges are 29.5--31.9\%, 48.9--50.0\%, and 67.6--67.9\% for the three targets. Thus, the budget account keeps the average workload near the selected target while allowing the coverage-derived demand to change the per-frame budget \(K_t\).

We next vary the switching threshold \(\theta_{iou}\) from 0.020 to 0.050. At \(r_{\mathrm{tar}}=30\%\), the success rate stays between 96.95\% and 97.10\%, while latency increases from 80.00 to 80.70\,ms. At \(r_{\mathrm{tar}}=70\%\), the success rate ranges from 97.45\% to 97.80\%, and latency changes by only 0.10\,ms. A larger \(\theta_{iou}\) makes Conservative Mode more likely because the condition \(\operatorname{IoU}_t\le\theta_{iou}\) holds more often. Within the tested range, however, its effect on success rate and latency is small. The two parameters therefore have different roles: \(r_{\mathrm{tar}}\) sets the main efficiency level, while \(\theta_{iou}\) adjusts the sensitivity of coverage-mode switching without substantially changing the average workload.

\begin{figure*}[h]
    \centering
    \includegraphics[width=\textwidth]{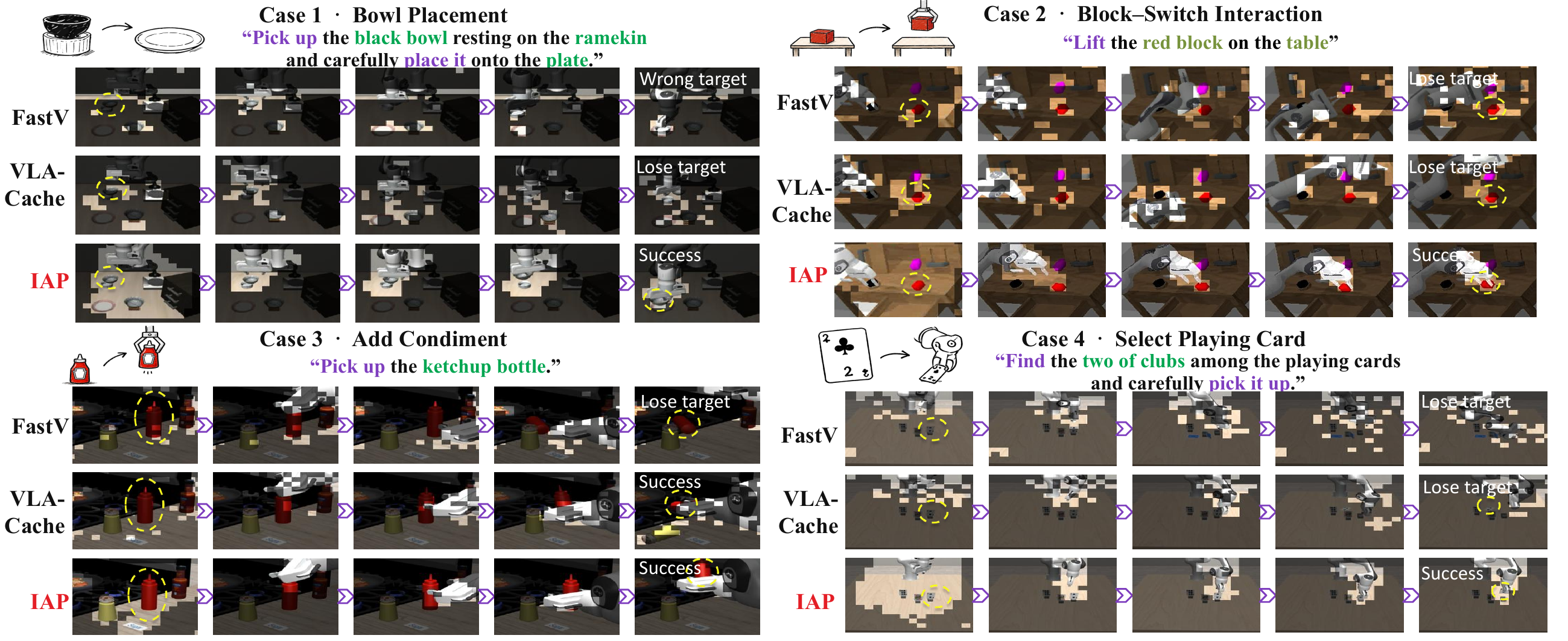}
     \caption{\textbf{Qualitative visual comparison under an extreme target average retention of 12.5\%.} Across four manipulation cases, IAprune preserves more continuous target- and manipulator-adjacent regions, while FastV and VLA-Cache tend to miss the target or keep fragmented redundant patches.}

    \label{fig:qualitative_visual_compare}
\end{figure*}
\begin{table}[!t]
\centering
\footnotesize
\renewcommand{\arraystretch}{1.15}
\setlength{\tabcolsep}{2.2pt}

\caption{\textbf{Budget tracking and threshold sensitivity on OpenVLA-OFT/LIBERO.}
The upper table reports observed average retention together with retention and latency ranges across the evaluated IoU thresholds. The lower table reports latency in milliseconds and success rate in percent as \(\theta_{iou}\) varies.}

\label{tab:robustness_scheduler}

\vspace{-1mm}

% ---------------- Average Budget Tracking ----------------
\begin{tabular*}{\columnwidth}{@{\extracolsep{\fill}}ccccc@{}}
\toprule[0.9pt]
\rowcolor{gray!8}
\multicolumn{5}{c}{\textbf{Average-Budget Tracking}} \\
\midrule[0.6pt]
\textbf{Target} 
& \textbf{Observed} 
& \textbf{Dev. (pp)} 
& \textbf{Ret. Range} 
& \textbf{Lat. Range (ms)} \\
\midrule[0.5pt]
30\% & \textbf{30.9\%} & \(\mathbf{+0.9}\) & 29.5--31.9\% & 78.2--82.6 \\
50\% & \textbf{49.7\%} & \(\mathbf{-0.3}\) & 48.9--50.0\% & 89.2--91.7 \\
70\% & \textbf{67.7\%} & \(\mathbf{-2.3}\) & 67.6--67.9\% & 98.5--99.9 \\
\bottomrule[0.9pt]
\end{tabular*}

\vspace{2.2mm}

% ---------------- IoU Threshold Trade-off ----------------
\begin{tabular*}{\columnwidth}{@{\extracolsep{\fill}}lccccccc@{}}
\toprule[0.9pt]
\rowcolor{gray!8}
\multicolumn{8}{c}{\textbf{IoU Threshold Trade-off}} \\
\midrule[0.6pt]
\textbf{\(\theta_{\mathrm{iou}}\)}
& \textbf{.020} & \textbf{.025} & \textbf{.030} & \textbf{.035}
& \textbf{.040} & \textbf{.045} & \textbf{.050} \\
\midrule[0.5pt]
T30 Lat. & \textbf{80.00} & 80.07 & 80.13 & 80.20 & 80.37 & 80.53 & 80.70 \\
T30 SR   & \textbf{97.10} & 97.00 & 96.95 & 97.00 & 97.05 & 97.00 & 97.00 \\
\midrule[0.5pt]
T70 Lat. & 98.50 & 98.52 & 98.54 & 98.56 & 98.57 & 98.58 & \textbf{98.60} \\
T70 SR   & 97.45 & 97.55 & 97.50 & 97.65 & 97.60 & 97.75 & \textbf{97.80} \\
\bottomrule[0.9pt]
\end{tabular*}

\vspace{-2mm}
\end{table}

\subsection{Qualitative Visualization}

Fig.~\ref{fig:qualitative_visual_compare} provides a closer look at token selection under a target average retention of 12.5\%. For IAprune, this value defines the average operating point rather than a fixed token count for every frame. In the four examples, IAprune retains more spatially connected regions around the target and manipulator. FastV and VLA-Cache more often omit target-adjacent tokens or retain fragmented patches. This pattern is consistent with the two decisions in IAprune: semantic--motion agreement selects the coverage mode, and the resulting coverage-derived demand guides \(K_t\); continuous semantic--motion relevance, followed by geometric residual correction, determines which existing tokens are retained within that budget. The controlled analyses above evaluate these components separately.

\subsection{Efficiency Analysis}

\subsubsection{Runtime and Memory}

\begin{table}[!t]
\centering
\renewcommand{\arraystretch}{1.00}
\setlength{\tabcolsep}{1.0pt}
\caption{\textbf{Memory and runtime on LIBERO.} Peak GPU memory and per-step CUDA runtime at nominal retention labels of 70\%, 50\%, and 30\%. For IAprune, these labels denote the target average retention ratio \(r_{\mathrm{tar}}\); baseline labels follow their original budget definitions.}

\label{tab:memory_runtime}
\begin{tabular*}{\columnwidth}{@{\extracolsep{\fill}}lcccccc@{}}
\toprule[0.9pt]
\multirow{2}{*}{\textbf{Method}} 
& \multicolumn{2}{c}{\textbf{70\% retention}} 
& \multicolumn{2}{c}{\textbf{50\% retention}} 
& \multicolumn{2}{c}{\textbf{30\% retention}} \\ 
\cmidrule(lr){2-3}\cmidrule(lr){4-5}\cmidrule(lr){6-7}
& \textbf{Mem.} & \textbf{Time} 
& \textbf{Mem.} & \textbf{Time} 
& \textbf{Mem.} & \textbf{Time} \\ 
\multicolumn{7}{r@{}}{Memory in GB; CUDA time in ms; lower is better.}\\[-1pt]
\midrule[0.7pt]

\multicolumn{7}{c}{\cellcolor[gray]{0.9}\textbf{DreamVLA on LIBERO}} \\ 
\midrule[0.5pt]
Vanilla (100\%) 
& \multicolumn{6}{c}{Unpruned: 2.661 GB memory / 101.47 ms CUDA time} \\ 
\midrule[0.5pt]
FastV 
& 2.626 & 84.55 
& 2.615 & 76.87 
& 2.608 & 72.96 \\
SparseVLM 
& 2.621 & 83.85 
& 2.612 & 75.16 
& 2.602 & 73.76 \\
DivPrune 
& 2.615 & 82.49 
& 2.594 & 75.72 
& 2.586 & 72.45 \\
VLA-Cache 
& 2.634 & 85.99 
& 2.628 & 84.56 
& 2.621 & 80.14 \\
\textbf{IAprune (Ours)} 
& \textbf{2.610} & \textbf{81.89} 
& \textbf{2.582} & \textbf{74.61} 
& \textbf{2.574} & \textbf{69.97} \\ 
\midrule[0.7pt]

\multicolumn{7}{c}{\cellcolor[gray]{0.9}\textbf{\(\pi_0\) on LIBERO}} \\ 
\midrule[0.5pt]
Vanilla (100\%) 
& \multicolumn{6}{c}{Unpruned: 6.214 GB memory / 94.13 ms CUDA time} \\ 
\midrule[0.5pt]
FastV 
& 6.135 & 79.44 
& 6.092 & 71.31 
& 6.079 & 66.76 \\
SparseVLM 
& 6.117 & 79.80 
& 6.082 & 70.83 
& 6.074 & 67.72 \\
DivPrune 
& 6.109 & 76.53 
& 6.076 & 71.85 
& 6.052 & 67.24 \\
VLA-Cache 
& 6.185 & 84.85 
& 6.147 & 80.45 
& 6.141 & 76.53 \\
\textbf{IAprune (Ours)} 
& \textbf{6.098} & \textbf{75.44} 
& \textbf{6.069} & \textbf{69.73} 
& \textbf{6.032} & \textbf{64.82} \\ 
\bottomrule[0.9pt]

\end{tabular*}
\end{table}
\vspace{-5pt}

Table~\ref{tab:memory_runtime} reports peak GPU memory and per-step CUDA runtime on LIBERO. For DreamVLA, the unpruned policy uses 2.661\,GB and 101.47\,ms. IAprune reduces the runtime to 81.89, 74.61, and 69.97\,ms at the 70\%, 50\%, and 30\% nominal retention labels, respectively; its peak memory reaches 2.574\,GB at 30\%. For \(\pi_0\), the unpruned policy uses 6.214\,GB and 94.13\,ms. IAprune reduces the runtime to 75.44, 69.73, and 64.82\,ms across the same labels, with peak memory reaching 6.032\,GB at 30\%. Among the compared methods, IAprune has the lowest runtime and peak memory at every reported retention label for both policies. These measurements show consistent compute and memory reductions in the two evaluated settings.

\begin{figure}[!t]
    \centering
    \includegraphics[width=0.9\columnwidth]{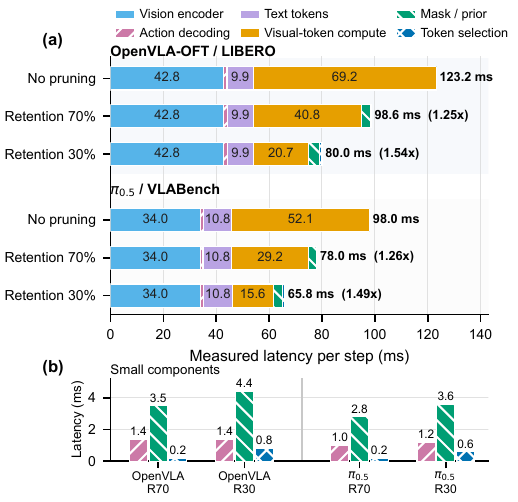}
      \caption{\textbf{Latency breakdown of IAprune.} Visual-token processing in the multimodal backbone is the main component that decreases with token retention. In the measured setting, vision encoding, text-token processing, and action decoding remain nearly constant.}

    \label{fig:latency_breakdown}
\end{figure}

\begin{figure*}[!t]
    \centering
    \includegraphics[width=\textwidth]{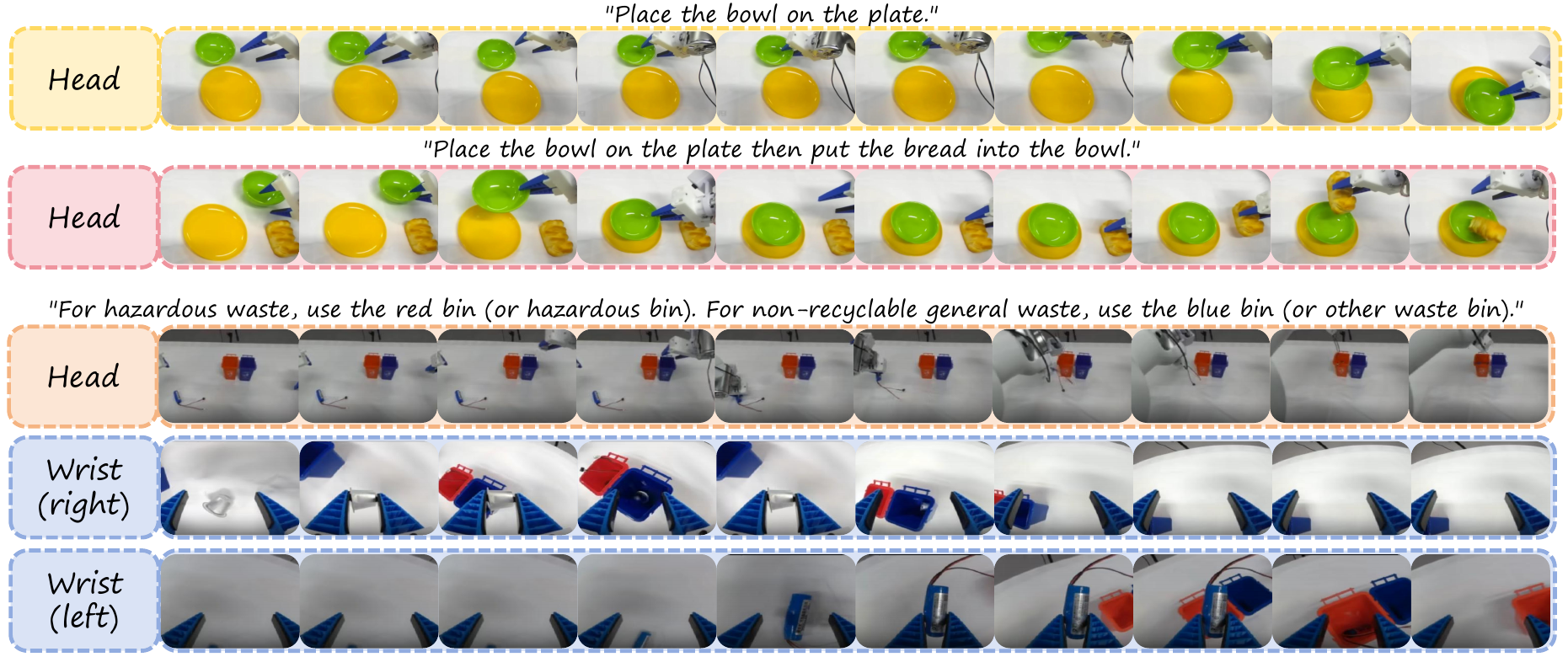}
    
    \caption{\textbf{Real-World Task Demonstrations.} From top to bottom: simple (single-step), long-horizon (multi-step), and dual-arm (coordinated) manipulation tasks, showing head and wrist camera perspectives.}
    \label{fig:real_world}
\end{figure*}

\subsubsection{Latency Breakdown}

Fig.~\ref{fig:latency_breakdown} separates the measured end-to-end latency into its main components. Panel~(a) shows that IAprune mainly reduces visual-token computation in the multimodal backbone, while vision encoding, text-token processing, and action decoding change little. Following Amdahl's law~\cite{amdahl1967validity}, let \(p=T_{\mathrm{vis}}/T_0\) be the fraction of the unpruned latency \(T_0\) spent on prunable visual-token computation. Let \(s\) denote the speedup of this component after pruning, and let \(\epsilon=T_{\mathrm{prune}}/T_0\) denote the normalized mask/prior construction and token-selection overhead. With the remaining costs treated as constant, the end-to-end speedup is approximated by

\begin{equation}
S_{\mathrm{e2e}}
\approx
\frac{1}{(1-p)+p/s+\epsilon}.
\label{eq:amdahl_latency}
\end{equation}
Even if visual-token computation were eliminated and pruning added no overhead, the unchanged fraction would impose an upper bound of \(S_{\max}=1/(1-p)\). In Fig.~\ref{fig:latency_breakdown}, visual-token computation accounts for \(p=69.2/123.2=0.562\) of the OpenVLA-OFT baseline and \(p=52.1/98.0=0.532\) of the \(\pi_{0.5}\) baseline. The corresponding ideal upper bounds are \(2.28\times\) and \(2.14\times\). Panel~(b) shows that mask/prior construction takes 2.8--4.4\,ms, while token selection takes 0.2--0.8\,ms. The reduction in visual-token computation is larger than this added overhead, resulting in a net decrease in measured latency. Eq.~\eqref{eq:amdahl_latency} also explains why end-to-end speedup remains bounded instead of scaling directly with the reduction in retained visual tokens.

\subsection{Real-Robot Evaluation}

Table~\ref{tab:real_world_experiments} reports success rate and inference latency on the three real-robot tasks. We run 25 trials for each method--task pair, and all pruning methods use the nominal 30\% retention setting. For IAprune, this setting denotes the target average retention ratio \(r_{\mathrm{tar}}=30\%\). Compared with the unpruned \(\pi_{0.5}\) policy, IAprune reduces single-arm latency from 88.1 to 59.7\,ms, corresponding to a $1.48\times$ speedup. On the dual-arm task, latency decreases from 124.3 to 84.6\,ms, corresponding to a $1.47\times$ speedup. The success rate increases from 80.0\% to 84.0\% on the Simple task and from 64.0\% to 68.0\% on the Long task. It remains 44.0\% on the Dual-arm task. The average success rate therefore changes from 62.7\% to 65.3\%. Fig.~\ref{fig:real_world} shows successful execution examples for the three tasks from the head and wrist RGB views.

\begin{table}[!t]
\centering
\scriptsize
\renewcommand{\arraystretch}{0.95}
\setlength{\tabcolsep}{2.5pt}

\caption{\textbf{Real-robot results on three tasks.} We report success rate and inference latency for the Simple, Long, and Dual-arm tasks. Vanilla denotes the unpruned \(\pi_{0.5}\) policy, and values in parentheses are speedups relative to Vanilla.}

\label{tab:real_world_experiments}

\resizebox{\columnwidth}{!}{%
\begin{tabular}{l cccc cc}
\toprule
\multirow{2}{*}{\textbf{Method}} & \multicolumn{4}{c}{\textbf{Success Rate (\%)}} & \multicolumn{2}{c}{\textbf{Latency (ms)}$\downarrow$} \\
\cmidrule(lr){2-5} \cmidrule(lr){6-7}
 & Simple & Long & Dual-Arm & \textbf{Avg.} & \textbf{Single} & \textbf{Dual} \\
\midrule
Vanilla (100\%) & 80.0 & 64.0 & 44.0 & 62.7 & 88.1 & 124.3 \\
\hline
FastV & 60.0 & 44.0 & 20.0 & 41.3 & 61.6 & 88.2 \\
VLA-ADP & 72.0 & 52.0 & 32.0 & 52.0 & 65.3 & 94.2 \\
\textbf{IAprune} & \textbf{84.0} & \textbf{68.0} & \textbf{44.0} & \textbf{65.3} & \textbf{59.7}\,\textcolor{red}{(1.48$\times$)} & \textbf{84.6}\,\textcolor{red}{(1.47$\times$)} \\
\bottomrule
\end{tabular}}
\end{table}

\section{Conclusion}
In this paper, we presented IAprune, a training-free method for reducing visual-token computation in embodied manipulation. IAprune treats pruning as two linked image-processing decisions. Semantic--motion agreement selects the coverage mode and guides the dynamic per-frame budget, while continuous relevance and geometric residual correction determine which existing tokens are retained without increasing that budget. Across four embodied manipulation policies, three simulation benchmarks, and a real-robot platform, IAprune matches the unpruned OpenVLA-OFT policy's 97.1\% average success rate at the target 30\% setting with a $1.54\times$ speedup, and provides up to a $1.48\times$ speedup on the real robot. Controlled analyses suggest that alignment-guided scheduling is most useful early in an episode when the token budget is tight, while structural reallocation redirects fixed selection slots toward under-represented boundaries. The current geometry branch relies on calibrated global and wrist views; extending it to less constrained camera settings remains future work.

\bibliographystyle{IEEEtran}
\bibliography{main}

\begin{IEEEbiography}[{\includegraphics[width=1in,height=1.25in,clip,keepaspectratio]{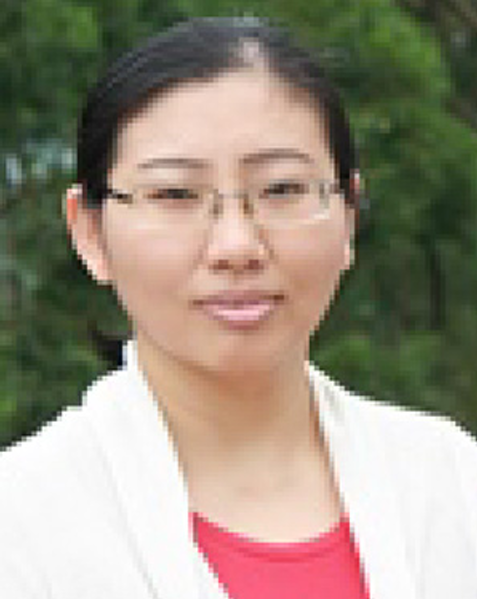}}]{Wei Zhang}
(Fellow, IEEE) received the Ph.D. degree in electrical engineering from Princeton University, Princeton, NJ, USA, in 2009, with the Wu Prize for Research Excellence. She is currently a professor with the Department of Electronic and Computer Engineering, The Hong Kong University of Science and Technology, Hong Kong, where she established the Reconfigurable System Lab. She was an Assistant Professor with the School of Computer Engineering, Nanyang Technological University, Singapore, from 2010 to 2013. She has authored and coauthored more than 150 papers in peer-reviewed journals and international conferences. Her current research interests include reconfigurable computing and FPGA-based design, electronic design automation, computer architecture, and embedded system security. Prof. Zhang won the Best Paper Awards at ISVLSI 2009, ICCAD 2017, and ICCAD 2022. She has served and currently serves on many editorial boards as an Associate Editor, including ACM Transactions on Reconfigurable Technology and Systems, IEEE Transactions on Computer-Aided Design of Integrated Circuits and Systems, ACM Transactions on Embedded Computing Systems, ACM Journal on Emerging Technologies in Computing Systems, IEEE Embedded Systems Letters, and IEEE Transactions on Circuits and Systems II: Express Briefs.
\end{IEEEbiography}

\end{document}